%% file: sn-article.tex
\documentclass[pdflatex,sn-mathphys-num]{sn-jnl}

\input{config/packages}
\input{config/commands}
\input{config/colors}
\input{config/figure-commands}

\input{config/theorems}


\begin{document}

\input{frontmatter/title-authors}
\input{frontmatter/abstract-keywords}
\maketitle

\input{sections/01-introduction}
\input{sections/02-related-work}
\input{sections/03-method}
\input{sections/04-experiments}
\input{sections/05-conclusion}

\appendix
\input{appendices/experiment-details}

\backmatter
\input{backmatter/declarations}

\bibliography{mosaic}

\end{document}

%% file: config/packages.tex
\usepackage{graphicx}
\usepackage{multirow}
\usepackage{amsmath,amssymb,amsfonts}
\usepackage{amsthm}
\usepackage{mathrsfs}
\usepackage[title]{appendix}
\usepackage{xcolor}
\usepackage{colortbl}
\usepackage{textcomp}
\usepackage{manyfoot}
\usepackage{booktabs}
\usepackage{tabularx}
\usepackage[shortlabels,inline]{enumitem}
\usepackage{xspace}
\usepackage{placeins}
\usepackage{tikz}
\usetikzlibrary{positioning,matrix,fit,chains,calc,backgrounds,
  arrows.meta,shapes.geometric,shapes.misc,decorations.pathreplacing}

%% file: config/commands.tex
\newcommand{\mosaic}{MoSAIC\xspace}

%% file: config/colors.tex
\definecolor{gold}{RGB}{255,215,0}
\definecolor{contentLite}{HTML}{DBEAFE}
\definecolor{contentDark}{HTML}{1D4ED8}
\definecolor{styleLite}{HTML}{FFEDD5}
\definecolor{styleDark}{HTML}{C2410C}
\definecolor{trajLite}{HTML}{DCFCE7}
\definecolor{trajDark}{HTML}{15803D}
\definecolor{denoiserLite}{HTML}{FEE2E2}
\definecolor{denoiserDark}{HTML}{B91C1C}
\definecolor{decoderLite}{HTML}{F3E8FF}
\definecolor{decoderDark}{HTML}{7E22CE}
\definecolor{ioLite}{HTML}{F3F4F6}
\definecolor{outputLite}{HTML}{FEF3C7}
\definecolor{outputDark}{HTML}{B45309}
\definecolor{ruleGray}{HTML}{6B7280}
\definecolor{zoneEnc}{HTML}{F8FAFC}
\definecolor{zoneDec}{HTML}{FAFAF9}
\definecolor{zoneEdge}{HTML}{E5E7EB}
\definecolor{partSpine}{HTML}{3B82F6}
\definecolor{partHead}{HTML}{0EA5E9}
\definecolor{partLArm}{HTML}{10B981}
\definecolor{partRArm}{HTML}{F59E0B}
\definecolor{partLLeg}{HTML}{F97316}
\definecolor{partRLeg}{HTML}{EF4444}

\definecolor{msInk}{HTML}{263238}
\definecolor{msMuted}{HTML}{66737F}
\definecolor{msRule}{HTML}{CBD2D9}
\definecolor{msPaper}{HTML}{FCFBF8}
\definecolor{msFrozen}{HTML}{F1F3F5}
\definecolor{msContent}{HTML}{3F6F9F}
\definecolor{msContentFill}{HTML}{EAF1F7}
\definecolor{msStyle}{HTML}{C46E59}
\definecolor{msStyleFill}{HTML}{F9ECE8}
\definecolor{msTraj}{HTML}{4E8A83}
\definecolor{msTrajFill}{HTML}{E8F3F1}
\definecolor{msProposed}{HTML}{5C63B5}
\definecolor{msProposedFill}{HTML}{EEEDFA}
\definecolor{msAux}{HTML}{A17D35}
\definecolor{msAuxFill}{HTML}{FBF3DF}
\definecolor{msTrain}{HTML}{756486}
\definecolor{msTrainFill}{HTML}{F3EFF7}
\definecolor{msSpine}{HTML}{7A5195}
\definecolor{msHead}{HTML}{D5A93B}
\definecolor{msLArm}{HTML}{4C78A8}
\definecolor{msRArm}{HTML}{72AFA9}
\definecolor{msLLeg}{HTML}{F28E2B}
\definecolor{msRLeg}{HTML}{E15759}

%% file: config/figure-commands.tex
\newcommand{\msgraphic}[2]{%
  \IfFileExists{#1}{\includegraphics[width=#2]{#1}}{%
    \begingroup\setlength{\fboxsep}{0pt}\color{msRule}%
    \fbox{\rule{0pt}{0.82cm}\rule{#2}{0pt}}\endgroup}}

%% file: config/theorems.tex
\theoremstyle{thmstyleone}

\theoremstyle{thmstyletwo}

\theoremstyle{thmstylethree}

%% file: frontmatter/title-authors.tex
\title[MoSAIC: Aligned Intervention Supervision for Part-Local Motion Style Transfer]
{MoSAIC: Aligned Intervention Supervision\\
for Part-Local Motion Style Transfer}

\author*[1]{\fnm{Nazanin} \sur{Amini}}
\email{Nazanin.Amini@utsa.edu}

\author[1]{\fnm{Kevin} \sur{Desai}}
\email{Kevin.Desai@utsa.edu}

\affil*[1]{%
  \orgdiv{Department of Computer Science, College of AI, Cyber and Computing},
  \orgname{The University of Texas at San Antonio},
  \orgaddress{
    \street{One UTSA Circle},
    \city{San Antonio},
    \state{Texas},
    \postcode{78249},
    \country{USA}
  }
}

%% file: frontmatter/abstract-keywords.tex
\abstract{Editing character motion often requires transferring a gesture or gait from one or more reference motions while preserving the source action, timing, root trajectory, and unselected body regions. Existing motion datasets, however, rarely provide paired targets for arbitrary part-local content--reference combinations, and self-reconstruction training may allow a diffusion model to reproduce the content motion while underusing the routed reference. We present MoSAIC, a latent diffusion framework for part-local reference-conditioned motion style transfer. MoSAIC factorizes content and reference features by anatomical region, preserves the root trajectory through a separate conditioning pathway, and routes user-selected references to individual body parts. Its central contribution is aligned intervention supervision, which constructs synchronized references and counterfactual targets through controlled local transformations, making both the requested regional response and the motion to be preserved directly observable during training. In a frozen evaluation comprising 128 motions and 896 routed conditions, part-masked routing reduces preserved-region error from 70.64 to 66.45~mm and matched-noise off-target leakage from 18.08 to 9.88~mm relative to whole-body routing, while retaining a positive selected-region response. A matched-budget continuation study further shows that retaining aligned intervention supervision produces an 8.8\% relative increase in selected-target response and a 2.0-percentage-point increase in requested-route influence concentration. These results demonstrate that MoSAIC improves the response--preservation trade-off required for selective and controllable part-local motion editing.
}

\keywords{Human motion style transfer, part-local motion editing, reference-conditioned motion generation, latent diffusion models, controllable character animation, aligned intervention supervision}

%% file: sections/01-introduction.tex
\section{Introduction}
\label{sec:intro}

Character-motion authoring often requires recombination rather than global
restylization. An animator may wish to preserve the action, timing, and root
trajectory of a content sequence while borrowing an arm gesture from one
reference and a leg gait from another. Such spatially selective reuse would
make captured motions more flexible assets for character animation, virtual
humans, and interactive visual-content creation. We study this capability as
\emph{part-local reference routing}: the user supplies a spatial source map
that assigns either the content motion or a reference to each anatomical
region. The assignment is fixed over the sequence and is not inferred by the
model. Figure~\ref{fig:mosaic_composition} illustrates the interface.

\input{figures/teaser/intro_teaser}

The central challenge is selective change rather than reference similarity
alone. A strong reference condition may create a visible local response while
also altering the action, root trajectory, contacts, or body regions that the
user did not select. A strong clean-content condition creates the opposite
failure mode: it may reproduce the source so faithfully that the local
reference has little effect. For example, transferring an arm gesture onto a
circular walk should neither straighten the path nor restyle the legs.
Part-local motion style transfer must therefore model both the requested
regional response and the motion that should remain unchanged.

Prior work establishes several components of this capability. Motion Puzzle
supports body-part style transfer and multi-reference
composition~\cite{jang2022motionpuzzle}; MoST uses separated content and style
encoders with part-attentive modulation~\cite{kim2024most}; and MCM-LDM
conditions latent diffusion on content, root trajectory, and a whole-body
motion reference~\cite{song2024mcmldm}. In text-guided editing, MotionReFit
constructs spatially blended training triplets~\cite{jiang2025motionrefit}.
These methods establish the value of anatomical conditioning and
multi-reference composition. MoSAIC focuses on a complementary question that
remains underexplored in reference-conditioned diffusion: how can training
directly specify the change induced by replacing one local reference while
simultaneously specifying the regions that should be preserved?

In self-reconstruction training, the content condition, style condition, and
denoising target all describe the same motion. A high-capacity clean-content
pathway can therefore explain the target while only weakly using the local
reference condition. Replacing one regional condition at inference then asks
the model to produce a response that was never explicitly observed during
training. Anatomical factorization and part-specific routing constrain
information flow, but they do not determine the intended effect of that
replacement or the complementary motion that should remain unchanged. We
refer to this unobserved local replacement response as the
\emph{missing-counterfactual supervision problem}. Aligned interventions turn
this underdetermined routing task into a supervised response--preservation
problem by making the intended regional change and preserved complement
observable during training.

\mosaic couples this formulation to a coordinated latent-diffusion
architecture (Fig.~\ref{fig:mosaic_architecture}). Part-factorized content and
style pathways expose the anatomical units to which sources are assigned,
whereas a separate root-trajectory condition retains global locomotion
information outside the style route. A local content bottleneck reduces the
clean-copying shortcut. The user-specified map selects a source for each
region, and part-specific modulation and temporal injection deliver the
selected cue to the evolving noisy sample. Anatomically constrained attention
restricts direct cross-part interaction while retaining skeletal
coordination. Together, these mechanisms translate the supervised local
response into controllable regional routing.

The central training mechanism is \emph{aligned intervention supervision}.
For a factual HumanML3D motion, a known kinematic transformation creates a
synchronized reference containing a controlled anatomical change. Applying
the same transformation only to the routed region produces an exact
counterfactual target. The factual motion, transformed reference, and target
share the underlying action, timing, sequence length, and root trajectory;
the exact changed and preserved regions are therefore known from the
constructed target, including legitimate kinematic propagation. This
same-sequence design provides a direct signal for selective anatomical control
without treating unrelated natural motions as frame-aligned or requiring
naturally paired local-style targets. Figure~\ref{fig:aligned_intervention}
illustrates the construction.

A frozen aligned counterfactual evaluation directly tests the resulting
response--preservation balance. Across 128 motions and 896 routed conditions,
masked routing achieves a fractional selected-target improvement of 0.198 over
the matched content-style control on position-identifiable conditions.
Relative to whole-body routing, it reduces preserved-region error by
4.19\,mm and matched-noise off-target leakage by 8.20\,mm. A matched-budget
continuation study further shows that retaining aligned intervention
supervision shifts the model toward stronger and more anatomically
concentrated use of routed references. Separate HumanML3D generation and CMU
whole-body evaluations verify that this local-control mechanism retains a
viable generation and whole-body transfer operating point.

Our contributions are threefold:
\begin{itemize}[leftmargin=*,itemsep=2pt,topsep=2pt]
  \item We formulate part-local reference-conditioned diffusion as a
        missing-counterfactual supervision problem, clarifying why
        self-reconstruction does not identify the response to a replaced local
        condition.
  \item We introduce aligned intervention supervision, which applies known
        local transformations to synchronized motions to construct exact
        training targets for both selected-region response and
        unselected-region preservation.
  \item We develop \mosaic, a part-routed latent diffusion model with
        region-specific content and reference pathways, a separate
        root-trajectory condition, anatomical attention, and user-specified
        single- or multi-reference source assignment, and evaluate it on a
        frozen protocol with matched-noise response, preservation, leakage,
        influence, and uncertainty analyses.
\end{itemize}

%% file: figures/teaser/intro_teaser.tex
% Figure 1: user-specified part-wise composition.
% The muted palette and typography are shared with the architecture figure.
\begin{figure*}[t]
\centering
\begin{tikzpicture}[
    font=\sffamily\fontsize{7.0}{7.9}\selectfont,
    panel/.style={
        draw=msRule, fill=white, line width=0.55pt,
        rounded corners=1.3pt, inner sep=0pt, outer sep=0pt
    },
    motionpanel/.style={panel},
    routepanel/.style={
        panel, draw=msProposed!58, line width=0.70pt
    },
    outputpanel/.style={
        panel, draw=msProposed!62, line width=0.76pt
    },
    stagehead/.style={
        align=center, text=msInk, inner sep=0pt,
        font=\sffamily\fontsize{7.3}{8.2}\selectfont\bfseries
    },
    cardhead/.style={
        text=msInk, inner sep=0pt,
        font=\sffamily\fontsize{7.2}{8.0}\selectfont\bfseries
    },
    descriptor/.style={
        text=msMuted, inner sep=0pt,
        font=\sffamily\fontsize{6.4}{7.1}\selectfont
    },
    tablehead/.style={
        text=msMuted, inner sep=0pt, minimum height=0.30cm,
        font=\sffamily\fontsize{6.3}{7.0}\selectfont\bfseries
    },
    routekey/.style={
        anchor=west, text=msInk, inner sep=0pt,
        minimum width=1.24cm, minimum height=0.40cm,
        font=\sffamily\fontsize{6.0}{6.8}\selectfont
    },
    sourcecell/.style={
        rounded corners=1.0pt, inner xsep=1.8pt, inner ysep=1.8pt,
        minimum width=1.17cm, minimum height=0.36cm, align=center,
        font=\sffamily\fontsize{6.3}{7.0}\selectfont\bfseries
    },
    flowarrow/.style={
        -{Stealth[length=3.1pt,width=3.6pt]},
        draw=msMuted, line width=0.74pt,
        shorten <=2.5pt, shorten >=3.0pt
    },
    composearrow/.style={
        flowarrow, draw=msProposed, line width=0.82pt
    }
]

% Reference identity uses two restrained hues.  These colors identify sources,
% not anatomical parts; anatomy is communicated by text and by the render.
\colorlet{figRefA}{msStyle}
\colorlet{figRefAFill}{msStyleFill}
\colorlet{figRefB}{msTrain}
\colorlet{figRefBFill}{msTrainFill}

% ---------------------------------------------------------------------------
% (a) INPUTS
% ---------------------------------------------------------------------------
\node[motionpanel, minimum width=3.55cm, minimum height=1.98cm]
    (content) at (0,1.025) {};
\node[motionpanel, minimum width=3.55cm, minimum height=1.91cm,
      below=0.14cm of content]
    (references) {};

\node[cardhead, anchor=north west, text=msContent]
    at ($(content.north west)+(0.13cm,-0.12cm)$) {Content motion};
\node[descriptor, anchor=south east]
    at ($(content.south east)+(-0.13cm,0.11cm)$) {circular walk};
\node[inner sep=0pt] at ($(content.center)+(0,-0.01cm)$)
    {\includegraphics[
        height=1.48cm,
        trim=60 163 168 27,
        clip
    ]{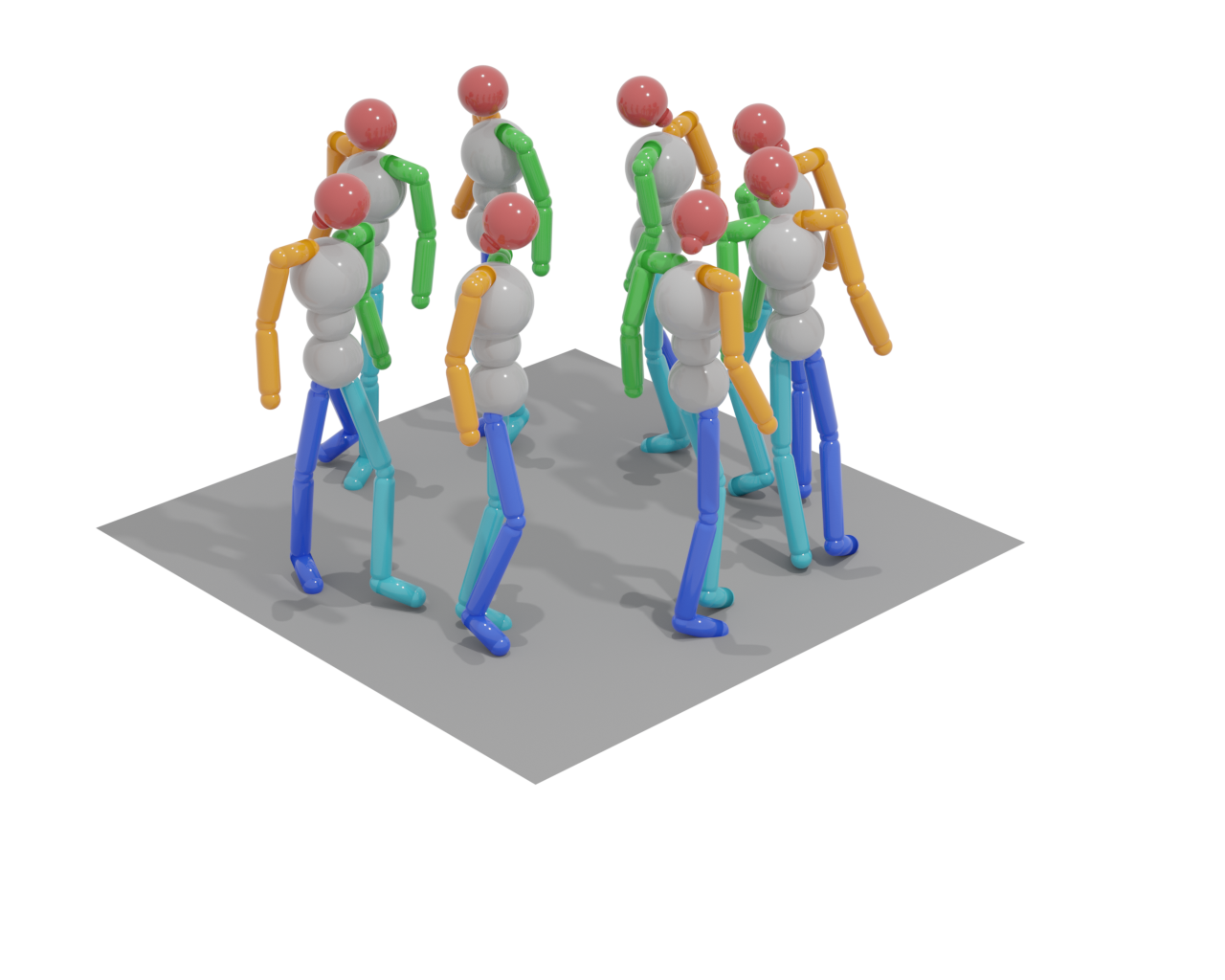}};
\draw[draw=msContent, line width=0.90pt]
    ($(content.north west)+(0.05cm,-0.03cm)$) --
    ($(content.north east)+(-0.05cm,-0.03cm)$);

\node[cardhead, anchor=north, text=figRefA]
    at ($(references.north)+(-0.88cm,-0.12cm)$) {Reference A};
\node[cardhead, anchor=north, text=figRefB]
    at ($(references.north)+(0.88cm,-0.12cm)$) {Reference B};
\node[inner sep=0pt] at ($(references.center)+(-0.88cm,-0.01cm)$)
    {\includegraphics[
        height=1.13cm,
        trim=203 220 179 12,
        clip
    ]{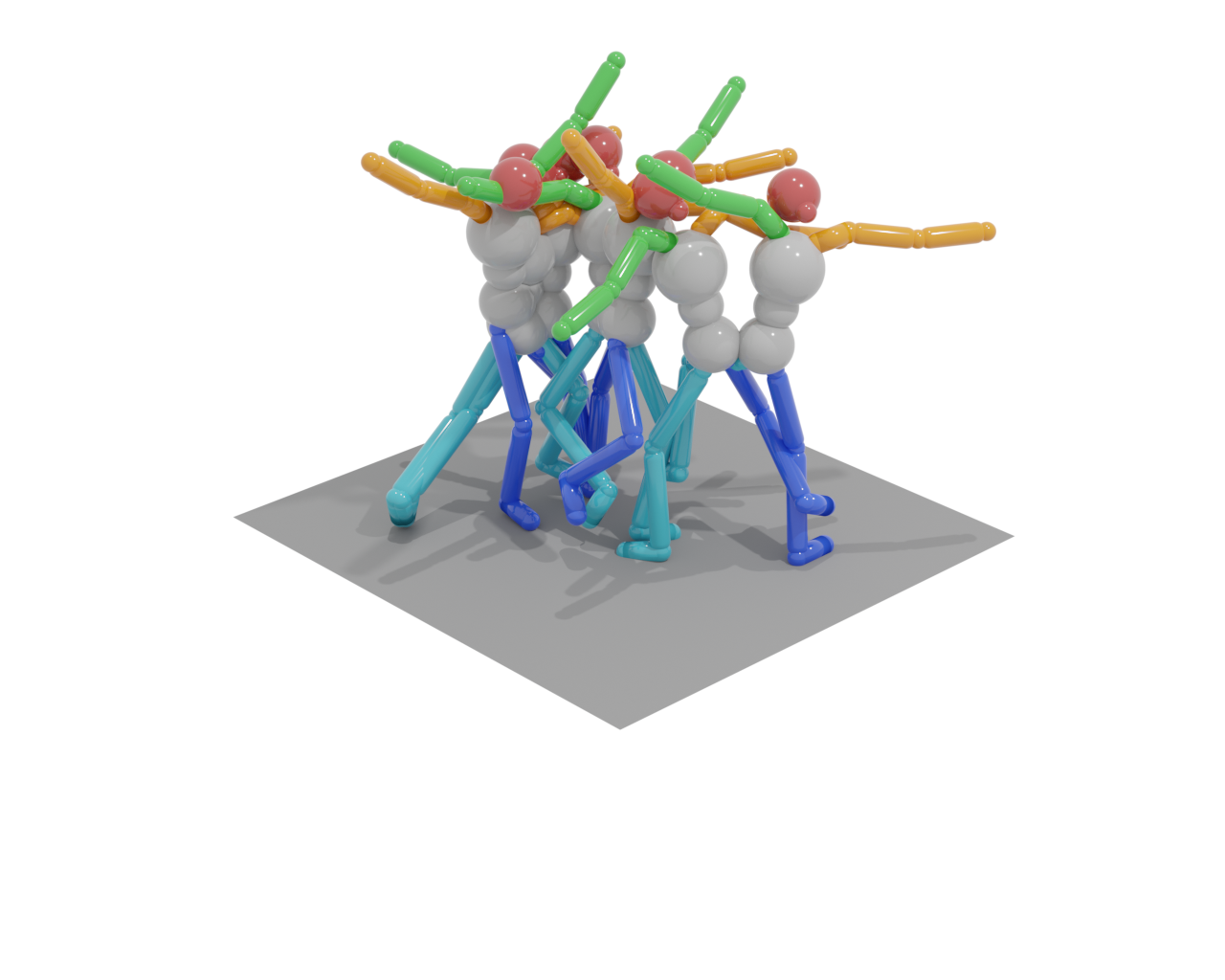}};
\node[inner sep=0pt] at ($(references.center)+(0.88cm,-0.01cm)$)
    {\includegraphics[
        height=1.13cm,
        trim=23 156 84 0,
        clip
    ]{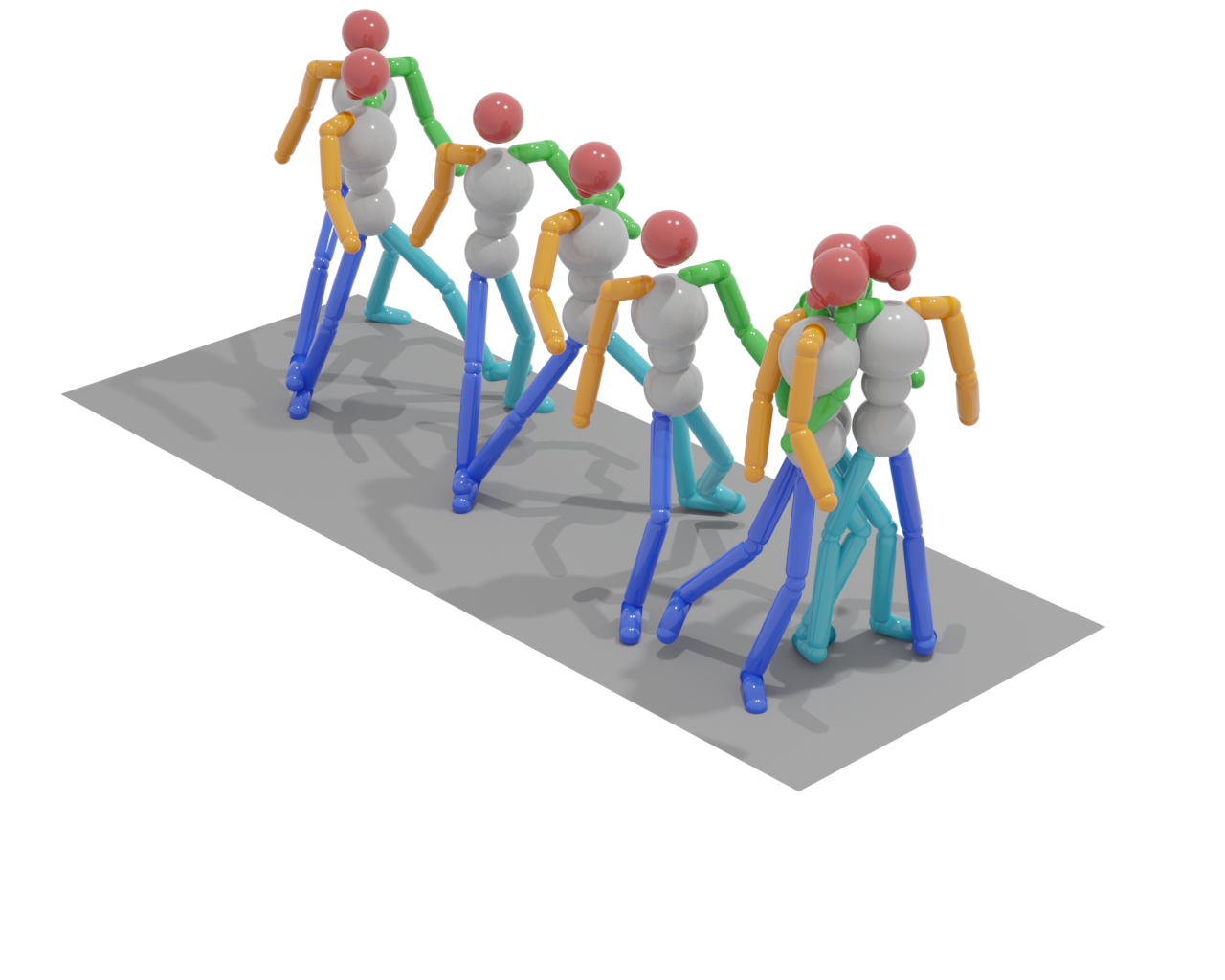}};
\node[descriptor, anchor=south, text=figRefA]
    at ($(references.south)+(-0.88cm,0.11cm)$) {hands high};
\node[descriptor, anchor=south, text=figRefB]
    at ($(references.south)+(0.88cm,0.11cm)$) {open legs};
\draw[draw=msRule!75, line width=0.42pt]
    ($(references.north)+(0,-0.28cm)$) --
    ($(references.south)+(0,0.19cm)$);
\draw[draw=figRefA, line width=0.90pt]
    ($(references.north west)+(0.05cm,-0.03cm)$) --
    ($(references.north)+(0,-0.03cm)$);
\draw[draw=figRefB, line width=0.90pt]
    ($(references.north)+(0,-0.03cm)$) --
    ($(references.north east)+(-0.05cm,-0.03cm)$);

\begin{scope}[on background layer]
    \node[fit=(content)(references), inner sep=0pt, outer sep=0pt]
        (inputGroup) {};
\end{scope}

% ---------------------------------------------------------------------------
% (b) USER-SPECIFIED ROUTING
% ---------------------------------------------------------------------------
\node[routepanel, minimum width=3.32cm, minimum height=4.03cm,
      right=0.58cm of inputGroup]
    (router) {};

\node[descriptor, anchor=north]
    at ($(router.north)+(0,-0.16cm)$) {fixed over the full sequence};

\matrix (routeTable) [
    matrix of nodes,
    ampersand replacement=\&,
    row sep=0.105cm,
    column sep=0.22cm,
    anchor=center
] at ($(router.center)+(0,-0.12cm)$) {
    |[tablehead, anchor=west]| Region
        \& |[tablehead, anchor=center]| Source \\
    |[routekey]| Arms
        \& |[sourcecell, draw=figRefA!55, fill=figRefAFill,
             text=figRefA]| Reference A \\
    |[routekey]| Spine + head
        \& |[sourcecell, draw=msContent!48, fill=msContentFill,
             text=msContent]| Content \\
    |[routekey]| Legs
        \& |[sourcecell, draw=figRefB!52, fill=figRefBFill,
             text=figRefB]| Reference B \\
    |[routekey]| Root trajectory
        \& |[sourcecell, draw=msTraj!50, fill=msTrajFill,
             text=msTraj]| Content \\
};

\draw[draw=msRule!78, line width=0.43pt]
    ($(routeTable-1-1.south west)+(0,-0.03cm)$) --
    ($(routeTable-1-2.south east)+(0,-0.03cm)$);
\draw[draw=msProposed, line width=0.94pt]
    ($(router.north west)+(0.05cm,-0.03cm)$) --
    ($(router.north east)+(-0.05cm,-0.03cm)$);

% ---------------------------------------------------------------------------
% (c) COMPOSED OUTPUT
% ---------------------------------------------------------------------------
\node[outputpanel, minimum width=4.56cm, minimum height=4.03cm,
      right=0.58cm of router]
    (output) {};

\node[descriptor, anchor=north]
    at ($(output.north)+(0,-0.16cm)$) {one denoising run};
\node[inner sep=0pt] at ($(output.center)+(0,0.06cm)$)
    {\includegraphics[
        height=2.72cm,
        trim=0 240 220 20,
        clip
    ]{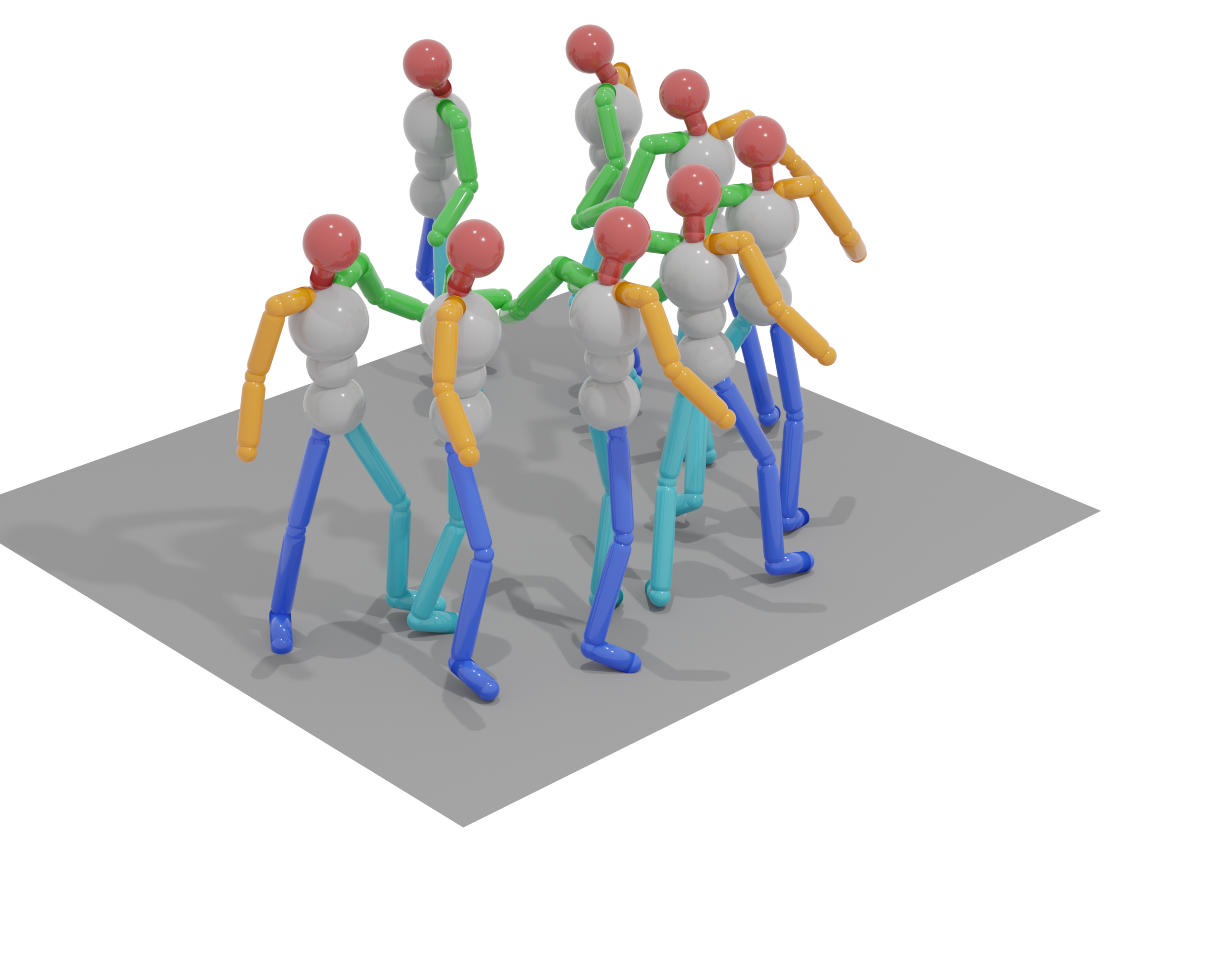}};
\node[descriptor, anchor=south, align=center]
    at ($(output.south)+(0,0.12cm)$)
    {\textcolor{figRefA}{A: arms}\enspace
     \textcolor{figRefB}{B: legs}\\[-0.2ex]
     \textcolor{msContent}{content: spine/head}\enspace
     \textcolor{msTraj}{content: trajectory}};
\draw[draw=msProposed, line width=0.98pt]
    ($(output.north west)+(0.05cm,-0.03cm)$) --
    ($(output.north east)+(-0.05cm,-0.03cm)$);
% Redraw the restrained output boundary after the opaque RGB render.
\draw[draw=msProposed!62, line width=0.76pt, rounded corners=1.3pt]
    (output.south west) rectangle (output.north east);

% ---------------------------------------------------------------------------
% STAGE LABELS AND FLOW
% ---------------------------------------------------------------------------
\node[stagehead, anchor=south] at ($(inputGroup.north)+(0,0.18cm)$)
    {(a) Input motions};
\node[stagehead, text width=3.32cm, anchor=south]
    at ($(router.north)+(0,0.18cm)$)
    {(b) User-specified routing};
\node[stagehead, anchor=south] at ($(output.north)+(0,0.18cm)$)
    {(c) MoSAIC output};

\draw[flowarrow] (inputGroup.east) -- (router.west);
\draw[composearrow] (router.east) -- (output.west);

\end{tikzpicture}
\caption{\textbf{User-specified part-wise composition.}
The content motion provides the action and root trajectory, while Reference~A
controls both arms and Reference~B controls both legs. MoSAIC combines the
routed conditions in a single denoising run at \(s=2.0\).}
\label{fig:mosaic_composition}
\end{figure*}

%% file: sections/02-related-work.tex
\section{Related Work}
\label{sec:related}

\subsection{Reference-Conditioned Motion Stylization}

Motion stylization seeks to change how an observed action is performed while
retaining its action structure and timing. Early example-based methods learned
correspondences or local dynamical mappings between performances
\cite{hsu2005style,xia2015realtime}; later neural systems used temporal
convolutions, feature statistics, or generative flows to separate and recombine
content and style
\cite{holden2016deep,holden2017fast,aberman2020unpaired,wen2021autoregressive}.
These formulations established the central response--preservation trade-off:
a reference must influence the output without overwriting the source motion
indiscriminately.

Recent work moves this problem into expressive latent and diffusion priors.
MotionCLIP aligns reconstructed motions with a semantic CLIP space and thereby
supports language-mediated generation and editing~\cite{tevet2022motionclip},
while MDM and MLD establish diffusion in pose and compressed motion spaces,
respectively~\cite{tevet2023mdm,chen2023mld}. For stylization specifically,
Generative Motion Stylization models deterministic content and probabilistic
style in a pretrained autoencoder space~\cite{guo2024genmostyle}, and MoST
combines Siamese encoders with part-attentive modulation across different
actions~\cite{kim2024most}. MCM-LDM extracts content, root trajectory, and
reference style as prioritized latent-diffusion conditions
\cite{song2024mcmldm}. SMooDi instead adapts a pretrained text-to-motion
diffusion model using a style adaptor and inference-time style guidance
\cite{zhong2024smoodi}. StyleMotif broadens the reference modality to motion,
text, image, video, and audio through style--content cross-fusion
\cite{guo2025stylemotif}, whereas MotionLab unifies generation, stylization,
and editing in a motion--condition--motion flow model
\cite{guo2025motionlab}. These methods provide strong global or
part-attentive reference conditioning. MoSAIC addresses the more specific
training problem of enforcing an explicit user-specified source assignment
independently for each anatomical region.

\subsection{Part-Aware Motion Transfer and Editing}

Motion Puzzle is the closest capability predecessor: skeleton-aware encoders
provide global and time-varying reference features to body-part adaptive
normalization and attention, and different references can be assigned to
different body regions~\cite{jang2022motionpuzzle}. MoST also uses body-part
attention, although its objective is transfer of a reference style across
diverse action content rather than execution of an explicit multi-source
route~\cite{kim2024most}. These studies establish anatomical factorization and
multi-reference composition as effective design principles. Relative to
Motion Puzzle, \mosaic changes both the generative model and the supervision:
six local content/style pathways and a separate trajectory condition enter a
latent diffusion transformer, while aligned counterfactual targets directly
supervise the intended routed effects.

Text-driven motion editing addresses related spatial control with a different
interface and evidence source. MotionReFit introduces MotionCutMix, which
forms source--instruction--target triplets by blending body parts from
different motions and coordinates the resulting edits
\cite{jiang2025motionrefit}. SimMotionEdit jointly learns motion-similarity
prediction and text-based editing to relate the requested change to source
preservation~\cite{li2025simmotionedit}. MotionMaster scales augmented
motion--language supervision in a multimodal language model and supports
generation, editing, and body-part composition
\cite{jiang2026motionmaster}. OmniME further makes change and invariance
explicit through positive--negative and motion-preservation objectives
\cite{shi2026omnime}. Such systems accept semantic instructions and may infer
where a requested edit applies. In contrast, \mosaic accepts motion references
and a spatial route supplied by the user. Its locality is anatomical rather
than strict independence in world coordinates, because a proximal rotation
can legitimately displace descendants through forward kinematics.

\subsection{Learning Selective Response and Preservation}

Content--style factorization alone does not identify the output that should
follow a local reference replacement. In general, unsupervised latent factors
are not uniquely identifiable without inductive bias or supervision
\cite{locatello2019challenging}. Reconstruction-based stylization can teach a
model to reproduce motions and feature statistics
\cite{aberman2020unpaired,jang2022motionpuzzle,kim2024most}, while architectural
separation of content, trajectory, and style can regulate their relative
influence~\cite{song2024mcmldm}. Neither signal supplies the missing
counterfactual: for the same factual motion, what should change when one
regional source is replaced, and what should remain invariant?

Editing datasets and objectives provide partial answers. MotionCutMix
constructs varied local edits from different source clips
\cite{jiang2025motionrefit}; MotionLab and SimMotionEdit learn paired
source-to-target transformations under semantic conditions
\cite{guo2025motionlab,li2025simmotionedit}; and OmniME explicitly penalizes
unwanted changes in text-guided editing~\cite{shi2026omnime}. These approaches
show that preservation is a supervised objective rather than an automatic
consequence of conditioning. Their targets, however, are designed for
instruction-based editing or combine motions whose timing and kinematics need
not be aligned.

Aligned intervention supervision targets the reference-routing case directly.
Starting from one HumanML3D training motion, a controlled local transformation
produces both a synchronized reference and a counterfactual target in which
only the routed anatomical region receives that intervention. Factual,
reference, and target sequences therefore share action, timing, length, and
root trajectory, making the requested response and the preserved complement
known during training. The transformations serve as controlled supervision for
learning selective response rather than as a taxonomy of natural motion
styles. Paired with an explicit six-part reference map, a part-local content
bottleneck, anatomical attention, and part-specific style injection, this
factual/counterfactual structure supports both training and matched evaluation
of response, preservation, and cross-part leakage.

%% file: sections/03-method.tex
\section{Method}
\label{sec:method}

\subsection{Problem Formulation and Overview}

Let $\mathbf{x}^{c}$ denote a content motion in the HumanML3D representation
\cite{guo2022generating}, and let
$\{\mathbf{x}^{s_k}\}_{k=1}^{K}$ contain one or two reference motions.  We use
six anatomical regions: spine, head, left and right arms, and left and right
legs.  A user supplies one source for each region:
\begin{equation}
 \mathcal{P}=\{\mathrm{spine},\mathrm{head},\mathrm{larm},
 \mathrm{rarm},\mathrm{lleg},\mathrm{rleg}\},\qquad
 \rho:\mathcal{P}\rightarrow\{0,\ldots,K\},\quad
 \mathbf{x}^{s_0}\equiv\mathbf{x}^{c}.
 \label{eq:routing}
\end{equation}
Source $0$ supplies content-derived local style, whereas sources $1$ and $2$
identify the references.  The assignment is spatial, externally specified,
and fixed throughout the sequence; MoSAIC neither predicts the route nor
localizes an edit in time.

The desired output $\widehat{\mathbf{x}}$ retains the content action,
temporal organization, and root trajectory while responding to the assigned
local reference cues. Unselected regions retain content-derived conditions.
Unlike whole-body transfer, the conditions may come from different motions
and must coexist without replacing their common content path. The problem
therefore requires both an anatomically indexed representation and supervision
that reveals which regional changes are requested and which complementary
motion should be preserved.
Figure~\ref{fig:mosaic_architecture} summarizes the pipeline.  A frozen motion
autoencoder factorizes content into local motion groups and a separate
trajectory condition.  Local reference encoders provide routed style
features, and a trainable diffusion transformer generates only the multipart
latent.  The frozen decoder then combines this latent with the retained
content trajectory.  Training uses aligned interventions for which both the
requested regional change and the complement to preserve are observable.
The inherited autoencoder supplies motion reconstruction, whereas routing,
content-capacity reduction, constrained conditioning, and aligned supervision
form the transfer mechanism studied here.

\input{figures/architecture/method_architecture}

\subsection{Part-Factorized Content and Style Conditioning}
\label{sec:method_conditions}

\noindent\textbf{Motion and trajectory.}
Following latent motion modeling \cite{chen2023mld}, a pretrained multipart
VAE encodes a normalized motion into six local latent groups and one
root-trajectory condition:
\begin{equation}
 E_{\mathrm{m}}(\mathbf{x})=
 \big(\mathbf{z}_1,\ldots,\mathbf{z}_6,\boldsymbol{\tau}\big),
 \qquad
 \mathbf{z}=[\mathbf{z}_1;\cdots;\mathbf{z}_6].
 \label{eq:part_latent}
\end{equation}
Only $\mathbf{z}$ is diffused; $\boldsymbol{\tau}$ remains a separate
condition. This separation keeps the content root trajectory outside the
routed style pathway.  The VAE encoder and decoder are frozen during diffusion
training.  Each local encoder receives the position, rotation, velocity, and,
where applicable, contact features associated with its region.  This is an
anatomical factorization rather than a claim of statistical independence:
shared joints and later denoiser communication still support coordination
between regions.

\noindent\textbf{Local style and content capacity.}
Each anatomical region has a temporal style encoder that reads only its local
non-root motion features and returns a pooled representation and an ordered
feature sequence.  Pooled features describe the overall local tendency,
whereas segment-pooled temporal features retain coarse phase order.  A small
semantic residual from a frozen MotionCLIP representation
\cite{tevet2022motionclip} supplements the pooled feature.  The local
extractors were pretrained without action, text, or style-category labels,
using local motion descriptors and intervention consistency as supervisory
signals.  During diffusion training, these extractors and the semantic
decomposer are frozen; their part-specific projections and bounded gates
remain trainable.  Thus the style pathway is partially, rather than entirely,
frozen, and root channels are excluded from its local inputs.

Directly exposing the complete clean latent would provide a strong copying
shortcut.  MoSAIC instead compresses each $\mathbf{z}_p$ independently into a
smaller clean-content memory $\mathbf{c}_{p,t}$ conditioned on the diffusion
timestep.  The trainable bottleneck reduces content capacity without mixing
anatomical regions or compressing the separate trajectory. By limiting the
clean-content shortcut, it encourages the denoiser to use routed reference
features while retaining region-specific content information. Applying the
same reducer independently to the six groups also preserves the correspondence
between each memory token and the region whose noisy latent it conditions.

Let $\mathbf{s}_p(\cdot)$ and $\mathbf{K}_p(\cdot)$ denote the projected
pooled and temporally ordered conditions.  Routing is deterministic selection
in condition space:
\begin{equation}
 \widetilde{\mathbf{s}}_p
 =\mathbf{s}_p(\mathbf{x}^{s_{\rho(p)}}),\qquad
 \widetilde{\mathbf{K}}_p
 =\mathbf{K}_p(\mathbf{x}^{s_{\rho(p)}}).
 \label{eq:source_routing}
\end{equation}
All regions not assigned to a reference therefore use the corresponding
content-derived condition rather than a zero vector.

\subsection{Part-Routed Diffusion Model}
\label{sec:method_denoiser}

Following epsilon-parameterized diffusion \cite{ho2020ddpm}, the concatenated
multipart latent is perturbed according to
\begin{equation}
 \mathbf{z}_t
 =\sqrt{\bar{\alpha}_t}\,\mathbf{z}
 +\sqrt{1-\bar{\alpha}_t}\,\boldsymbol{\epsilon},
 \qquad
 \boldsymbol{\epsilon}\sim\mathcal{N}(\mathbf{0},\mathbf{I}).
 \label{eq:forward_diffusion}
\end{equation}
The DiT-style denoiser \cite{peebles2023scalable} receives the noisy latent
tokens after a clean prefix formed by the six local content memories, and
predicts noise only for the multipart latent.  Its attention structure gives
generated tokens one-way access to clean content: noisy tokens may read the
memory, but memory queries cannot read the evolving sample.  Within the noisy
sample, anatomical attention limits direct interaction to the same or
neighboring regions, including registered bilateral and torso connections.
This mask operates on sample-to-sample attention rather than on the clean
prefix.  Indirect coordination through anatomical neighbors and content memory remains
possible. The mask therefore restricts direct cross-region propagation while
preserving the coordination needed for coherent whole-body motion.

Three complementary signals condition each denoising block.  First, the
retained root trajectory modulates the feed-forward pathway independently of
style.  Second, every noisy token receives pooled style modulation from
$\widetilde{\mathbf{s}}_p$ for its anatomical region; reference style is not
applied to the clean prefix.  This pooled condition is available throughout
the denoiser and controls part-specific affine modulation and residual
strength.  Third, a constrained cross-attention path provides the ordered
keys $\widetilde{\mathbf{K}}_p$.  A noisy query may attend only to keys from
its own region, while clean-memory queries receive no temporal-style update.
This temporal path is gated, used in later denoising blocks, and restricted
to selected noise levels, where reference dynamics can be introduced without
directly rewriting content memory.  Together, content-memory isolation,
anatomical attention, and part-specific style injection limit direct
cross-region influence while retaining whole-body coordination.

\subsection{Aligned Intervention Supervision and Training Objective}
\label{sec:method_intervention}

Natural training motions do not provide the target for an arbitrary
part-to-reference route.  We therefore start from a factual training motion
$\mathbf{x}^{A}$ and apply a controlled local transformation
$\mathcal{T}$ to its canonical region set $C$, producing an aligned reference
\begin{equation}
 \mathbf{x}^{B}=\mathcal{T}_{C}(\mathbf{x}^{A}).
 \label{eq:aligned_reference}
\end{equation}
For a sampled routed set $M\subseteq\mathcal{P}$, the counterfactual target
applies the same intervention only where the route overlaps its canonical
support:
\begin{equation}
 \mathbf{x}^{M}=\mathcal{T}_{M\cap C}(\mathbf{x}^{A}),
 \label{eq:aligned_target}
\end{equation}
The factual motion, reference, and target share action, timing, sequence
length, and root trajectory.  The target therefore specifies both the
required local response and the complement to preserve.  The active
transformations cover controlled torso, head, elbow, and knee changes and
update dependent kinematics consistently.  A nontrivial route is sampled with
nonempty overlap between $M$ and $C$, ensuring that every intervention row
contains an observable requested change.  A proximal joint rotation may move
descendant positions through the kinematic chain; preservation is therefore
defined against the constructed target rather than by requiring every
off-route coordinate to remain identical.  The transformations therefore provide controlled supervision for selective
response while preserving the kinematic dependencies represented by the
counterfactual target.

\input{figures/results/aligned-intervention}

Content memory and trajectory always come from $A$.  Self rows denoise its
latent, whereas intervention rows denoise the target latent while receiving
the exact part-wise mixture of style conditions from $A$ and $B$.  All rows
use epsilon-prediction loss; self rows cover the full diffusion schedule.
Intervention rows use the configured lower-noise interval so that the
one-step clean estimate used for motion-space supervision remains meaningful.
Once the intervention curriculum is active, these rows additionally use
requested-region response, unselected-region preservation,
physical-consistency, and local style-code losses:
\begin{equation}
 \begin{aligned}
 \mathcal{L}_{\mathrm{diff}}
 &=\mathbb{E}\!\left[
 \|\boldsymbol{\epsilon}
 -\boldsymbol{\epsilon}_{\theta}(\mathbf{z}_t,t,\mathbf{c})\|_2^2\right],
 \\
 \mathcal{L}_{\mathrm{total}}
 &=\mathcal{L}_{\mathrm{diff}}+\kappa(e)\big(
 \lambda_{\mathrm{resp}}\mathcal{L}_{\mathrm{resp}}
 +\lambda_{\mathrm{pres}}\mathcal{L}_{\mathrm{pres}}
 +\lambda_{\mathrm{phys}}\mathcal{L}_{\mathrm{phys}}
 +\lambda_{\mathrm{style}}\mathcal{L}_{\mathrm{style}}\big).
 \end{aligned}
 \label{eq:total_loss}
\end{equation}
Here $\mathbf{c}$ contains the local content memory, trajectory, and routed
style conditions, and $\kappa(e)$ introduces the auxiliary terms gradually.
After a one-step clean estimate is decoded with the frozen decoder,
$\mathcal{L}_{\mathrm{resp}}$ matches routed feature columns to the aligned
reference and $\mathcal{L}_{\mathrm{pres}}$ matches the complement to
$\mathbf{x}^{M}$.  The physical group covers root trajectory, contact,
velocity, and motion-statistic consistency.  The style group re-encodes the
prediction and combines part-code matching, a source margin, and cross-part
decorrelation; its reference codes are stop-gradient targets.  The numerical
weights, curriculum, and intervention sampling schedule are fixed by the
released training configuration.  Style-condition dropout exposes the
denoiser to the zero-style branch used at inference.  The selected training path makes one denoising prediction per row rather than
adding a second factual--counterfactual branch.

\subsection{Inference and Multi-Reference Routing}

At inference, the frozen VAE mean-encodes the content motion, and the style
pathway encodes one or two references.  The user-supplied source map selects
pooled and temporal conditions for every region, with content-derived style
assigned to unselected regions.  DDIM \cite{song2021ddim} samples the
multipart latent while the content memory and root trajectory remain
identical in both classifier-free branches.  With zero-style prediction
$\boldsymbol{\epsilon}_{0}$ and routed prediction
$\boldsymbol{\epsilon}_{\rho}$, guidance is
\begin{equation}
 \widehat{\boldsymbol{\epsilon}}
 =\boldsymbol{\epsilon}_{0}
 +s\left(
 \boldsymbol{\epsilon}_{\rho}
 -\boldsymbol{\epsilon}_{0}\right),
 \label{eq:cfg}
\end{equation}
where the first branch receives zero pooled and temporal style conditions and
the second receives Eq.~\ref{eq:source_routing} \cite{ho2022cfg}.  The frozen
decoder combines the generated multipart latent with the unchanged content
trajectory.  Two references can thus condition different spatial regions in
one denoising run without retraining or independently sampling and stitching
regional outputs.  Routing remains externally specified and fixed over time;
the implementation does not support temporal route changes.

%% file: figures/architecture/method_architecture.tex
% Human-designed revision of the MoSAIC architecture figure.
% Required TikZ libraries: positioning, calc, fit, backgrounds, arrows.meta.
% The figure uses the existing ms* palette and \msgraphic helper.

\begin{figure}[t]
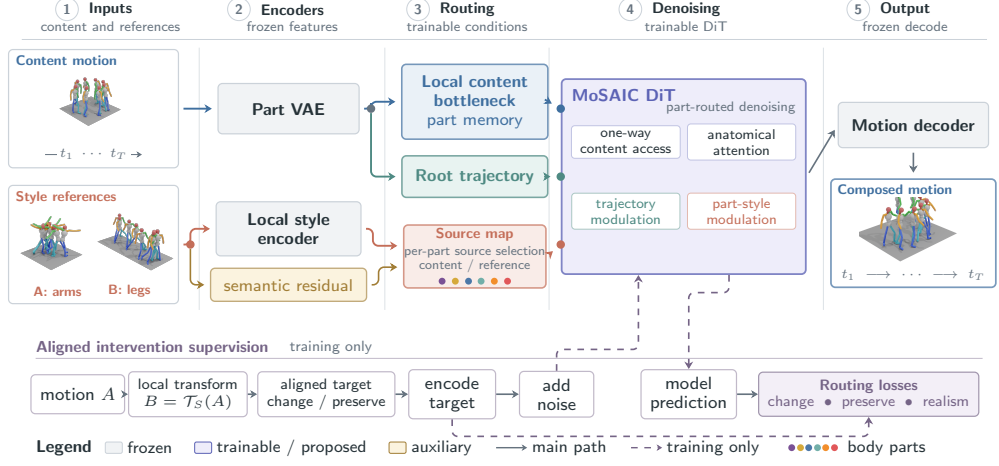

\centering
\resizebox{\textwidth}{!}{%
\begin{tikzpicture}[
    font=\sffamily\fontsize{7.7}{8.7}\selectfont,
    >={Stealth[length=3.4pt,width=4.0pt]},
    basecard/.style={
        draw=msRule, fill=white, line width=0.55pt,
        rounded corners=2.0pt, align=center,
        inner xsep=6pt, inner ysep=4.8pt
    },
    frozen/.style={basecard, fill=msFrozen, text=msInk},
    content/.style={
        basecard, draw=msContent!72, fill=msContentFill,
        text=msContent, line width=0.72pt
    },
    style/.style={
        basecard, draw=msStyle!72, fill=msStyleFill,
        text=msStyle, line width=0.72pt
    },
    trajectory/.style={
        basecard, draw=msTraj!75, fill=msTrajFill,
        text=msTraj, line width=0.72pt
    },
    proposed/.style={
        basecard, draw=msProposed!78, fill=msProposedFill,
        text=msProposed, line width=0.82pt
    },
    auxiliary/.style={
        basecard, draw=msAux!65, fill=msAuxFill,
        text=msAux, line width=0.62pt
    },
    motioncard/.style={
        basecard, minimum width=2.70cm, minimum height=1.88cm
    },
    outputcard/.style={
        basecard, draw=msContent!70, minimum width=2.62cm,
        minimum height=1.72cm, line width=0.78pt
    },
    stagehead/.style={
        anchor=west, inner sep=0pt, text=msInk,
        font=\sffamily\fontsize{6.9}{7.6}\selectfont\bfseries
    },
    stagesub/.style={
        anchor=west, inner sep=0pt, text=msMuted,
        font=\sffamily\fontsize{6.3}{6.9}\selectfont
    },
    stageindex/.style={
        circle, draw=msRule, fill=white, text=msMuted,
        minimum size=0.36cm, inner sep=0pt,
        font=\sffamily\fontsize{6.3}{6.8}\selectfont\bfseries
    },
    innerblock/.style={
        draw=msProposed!38, fill=white, rounded corners=1.5pt,
        minimum width=1.72cm, minimum height=0.50cm,
        inner xsep=4pt, inner ysep=2.4pt, align=center,
        text=msInk, font=\sffamily\fontsize{6.4}{7.1}\selectfont
    },
    traincard/.style={
        basecard, draw=msTrain!45, fill=white,
        minimum height=0.62cm, inner xsep=4.2pt, inner ysep=3.2pt,
        text=msInk, font=\sffamily\fontsize{6.8}{7.5}\selectfont
    },
    legendtext/.style={
        anchor=west, inner sep=0pt, text=msInk,
        font=\sffamily\fontsize{7.5}{8.2}\selectfont
    },
    swatch/.style={
        minimum width=0.27cm, minimum height=0.18cm,
        rounded corners=0.8pt, inner sep=0pt
    },
    port/.style={circle, minimum size=3.8pt, inner sep=0pt},
    partdot/.style={circle, minimum size=3.1pt, inner sep=0pt},
    flow/.style={
        -{Stealth[length=3.4pt,width=4.0pt]}, draw=msMuted,
        line width=0.74pt, rounded corners=2.5pt,
        shorten <=2pt, shorten >=2.5pt
    },
    contentflow/.style={flow, draw=msContent, line width=0.86pt},
    styleflow/.style={flow, draw=msStyle, line width=0.86pt},
    trajflow/.style={flow, draw=msTraj, line width=0.86pt},
    auxflow/.style={flow, draw=msAux, line width=0.72pt},
    trainflow/.style={
        -{Stealth[length=3.4pt,width=4.0pt]}, draw=msTrain,
        dashed, line width=0.74pt, rounded corners=2.5pt,
        shorten <=0pt, shorten >=0pt
    },
    stripflow/.style={
        -{Stealth[length=3.2pt,width=3.7pt]}, draw=msMuted,
        line width=0.74pt, rounded corners=2.0pt,
        shorten <=0pt, shorten >=0pt
    }
]

% ---------------------------------------------------------------------------
% Stage headers. Thin rules and open spacing replace heavy infographic boxes.
% ---------------------------------------------------------------------------
\foreach \x/\n/\title/\sub in {
    1.19/1/Inputs/{content and references},
    4.08/2/Encoders/{frozen features},
    6.86/3/Routing/{trainable conditions},
    10.34/4/Denoising/{trainable DiT},
    13.82/5/Output/{frozen decode}
}{
    \node[stagehead, anchor=center] (stageTitle\n) at (\x,2.66) {\title};
    \node[stageindex] at
        ($(stageTitle\n.west)+(-0.34cm,0)$) {\n};
    \node[stagesub, anchor=center] at (\x,2.39) {\sub};
}
\draw[msRule, line width=0.45pt] (-0.22,2.18) -- (15.12,2.18);
\foreach \x in {2.60,5.55,8.16,12.52}{
    \draw[msRule!65, line width=0.36pt] (\x,2.05) -- (\x,-2.12);
}

% ---------------------------------------------------------------------------
% Inputs
% ---------------------------------------------------------------------------
\node[motioncard] (contentInput) at (0.92,1.08) {};
\node[anchor=north west, inner sep=0pt,
      font=\sffamily\fontsize{6.3}{6.9}\selectfont\bfseries,
      text=msContent]
    at ($(contentInput.north west)+(0.12cm,-0.10cm)$) {Content motion};
\node at ($(contentInput.center)+(0,0.04cm)$)
    {\msgraphic{figures/teaser/content_walk_circle.png}{1.28cm}};
\draw[-{Stealth[length=2.5pt,width=2.9pt]}, draw=msMuted,
      line width=0.42pt]
    ($(contentInput.south)+(-0.77cm,0.22cm)$) --
    ($(contentInput.south)+(0.77cm,0.22cm)$);
\node[font=\sffamily\fontsize{6.1}{6.7}\selectfont,
      text=msMuted, fill=white, inner xsep=1.4pt, inner ysep=0pt]
    at ($(contentInput.south)+(0,0.23cm)$) {$t_1\;\cdots\;t_T$};

\node[motioncard] (styleInput) at (0.92,-1.08) {};
\node[anchor=north west, inner sep=0pt,
      font=\sffamily\fontsize{6.3}{6.9}\selectfont\bfseries,
      text=msStyle]
    at ($(styleInput.north west)+(0.12cm,-0.10cm)$) {Style references};
\node[inner sep=0pt] at ($(styleInput.center)+(-0.67cm,0.02cm)$)
    {\includegraphics[
        height=0.95cm,
        trim=203 220 179 12,
        clip
    ]{figures/teaser/style_hands_high.png}};
\node[inner sep=0pt] at ($(styleInput.center)+(0.67cm,0.02cm)$)
    {\includegraphics[
        height=0.95cm,
        trim=23 156 84 0,
        clip
    ]{figures/teaser/style_open_leg.png}};
\node[font=\sffamily\fontsize{6.1}{6.7}\selectfont\bfseries,
      text=msStyle, inner sep=0pt]
    at ($(styleInput.south)+(-0.57cm,0.21cm)$) {A: arms};
\node[font=\sffamily\fontsize{6.1}{6.7}\selectfont\bfseries,
      text=msStyle, inner sep=0pt]
    at ($(styleInput.south)+(0.57cm,0.21cm)$) {B: legs};

% ---------------------------------------------------------------------------
% Encoders
% ---------------------------------------------------------------------------
\node[frozen, minimum width=2.18cm, minimum height=0.80cm,
      text width=1.82cm] (vae) at (4.02,1.08)
    {\textbf{Part VAE}};

\node[frozen, minimum width=2.18cm, minimum height=0.86cm,
      text width=1.88cm] (styleEncoder) at (4.02,-0.83)
    {\textbf{Local style encoder}};

\node[auxiliary, minimum width=2.18cm, minimum height=0.50cm]
    (semanticResidual) at (4.02,-1.72)
    {semantic residual};

% ---------------------------------------------------------------------------
% Routing and conditioning
% ---------------------------------------------------------------------------
\node[content, minimum width=2.26cm, minimum height=0.94cm,
      text width=1.90cm] (contentBottleneck) at (6.98,1.19)
    {\textbf{Local content bottleneck}\\part memory};

\node[trajectory, minimum width=2.26cm, minimum height=0.66cm]
    (rootTrajectory) at (6.98,0)
    {\textbf{Root trajectory}};

\node[style, minimum width=2.26cm, minimum height=1.00cm]
    (sourceMap) at (6.98,-1.27) {};
\node[font=\sffamily\fontsize{6.5}{7.1}\selectfont\bfseries,
      text=msStyle, inner sep=0pt]
    at ($(sourceMap.north)+(0,-0.16cm)$) {Source map};
\node[font=\sffamily\fontsize{5.9}{6.5}\selectfont,
      text=msMuted, align=center, inner sep=0pt]
    at ($(sourceMap.center)+(0,-0.04cm)$)
    {per-part source selection\\content / reference};
\foreach \dx/\c in {-0.52/msSpine,-0.31/msHead,-0.10/msLArm,
                     0.10/msRArm,0.31/msLLeg,0.52/msRLeg}{
    \node[partdot, fill=\c] at ($(sourceMap.south)+(\dx cm,0.12cm)$) {};
}

% ---------------------------------------------------------------------------
% Part-routed diffusion transformer
% ---------------------------------------------------------------------------
\node[proposed, minimum width=3.88cm, minimum height=3.10cm]
    (dit) at (10.30,0) {};
\node[anchor=north west, inner sep=0pt,
      font=\sffamily\fontsize{7.8}{8.5}\selectfont\bfseries,
      text=msProposed]
    at ($(dit.north west)+(0.17cm,-0.16cm)$) {MoSAIC DiT};
\node[anchor=north east, inner sep=0pt,
      font=\sffamily\fontsize{6.2}{6.8}\selectfont,
      text=msMuted]
    at ($(dit.north east)+(-0.17cm,-0.36cm)$) {part-routed denoising};

\node[port, fill=msContent] (portContent) at ($(dit.west)+(0,1.08cm)$) {};
\node[port, fill=msTraj] (portTraj) at (dit.west) {};
\node[port, fill=msStyle] (portStyle) at ($(dit.west)+(0,-1.08cm)$) {};

\node[innerblock] (oneWay) at ($(dit.center)+(-0.92cm,0.55cm)$)
    {one-way\\content access};
\node[innerblock] (anatomy) at ($(dit.center)+(0.92cm,0.55cm)$)
    {anatomical\\attention};
\node[innerblock, draw=msTraj!48, text=msTraj]
    (trajMod) at ($(dit.center)+(-0.92cm,-0.55cm)$)
    {trajectory\\modulation};
\node[innerblock, draw=msStyle!48, text=msStyle]
    (styleMod) at ($(dit.center)+(0.92cm,-0.55cm)$)
    {part-style\\modulation};

% ---------------------------------------------------------------------------
% Decoder and output
% ---------------------------------------------------------------------------
\node[frozen, minimum width=2.16cm, minimum height=0.72cm]
    (decoder) at (13.95,0.82) {\textbf{Motion decoder}};

\node[outputcard] (output) at (13.95,-0.90) {};
\node at ($(output.center)+(0,0.03cm)$)
    {\msgraphic{figures/teaser/output_composed.png}{1.72cm}};
\node[anchor=north west, inner sep=0pt, fill=white,
      font=\sffamily\fontsize{6.25}{6.85}\selectfont\bfseries,
      text=msContent]
    at ($(output.north west)+(0.11cm,-0.09cm)$) {Composed motion};
\node[anchor=south, inner sep=0pt,
      font=\sffamily\fontsize{6.1}{6.7}\selectfont,
      text=msMuted]
    at ($(output.south)+(0,0.12cm)$)
    {$t_1\;\longrightarrow\;\cdots\;\longrightarrow\;t_T$};

% ---------------------------------------------------------------------------
% Main connectors. Three semantic lanes remain visually independent.
% ---------------------------------------------------------------------------
\coordinate (vaeFork) at ($(vae.east)+(0.18cm,0)$);
\coordinate (styleFork) at ($(styleInput.east)+(0.18cm,0)$);

\draw[contentflow] (contentInput.east) -- (vae.west);
\draw[contentflow] (vae.east) -- (vaeFork) |- (contentBottleneck.west);
\draw[trajflow] (vaeFork) |- (rootTrajectory.west);
\node[port, fill=msMuted] at (vaeFork) {};

\draw[styleflow] (styleInput.east) -- (styleFork) |- (styleEncoder.west);
\draw[auxflow] (styleFork) |- (semanticResidual.west);
\node[port, fill=msStyle] at (styleFork) {};
\draw[styleflow] (styleEncoder.east) -- ++(0.18cm,0) |-
    ($(sourceMap.west)+(0,0.17cm)$);
\draw[auxflow] (semanticResidual.east) -- ++(0.18cm,0) |-
    ($(sourceMap.west)+(0,-0.17cm)$);

\draw[contentflow] (contentBottleneck.east) -- (portContent);
\draw[trajflow] (rootTrajectory.east) -- (portTraj);
\draw[styleflow] (sourceMap.east) -- (portStyle);

\draw[flow] (dit.east) -- (decoder.west);
\draw[flow] (decoder.south) -- (output.north);

% ---------------------------------------------------------------------------
% Training-only strip. A compact linear construction and two vertical dashed
% links replace the previous long outside routing paths.
% ---------------------------------------------------------------------------
\node[anchor=west, inner sep=0pt,
      font=\sffamily\fontsize{7.2}{7.9}\selectfont\bfseries,
      text=msTrain] (trainingTitle) at (0.00,-2.72)
    {Aligned intervention supervision};
\node[anchor=west, inner sep=0pt,
      font=\sffamily\fontsize{6.7}{7.3}\selectfont,
      text=msMuted] at ($(trainingTitle.east)+(0.32cm,0)$) {training only};
\draw[msTrain!45, line width=0.42pt] (0.00,-2.89) -- (15.05,-2.89);

\node[traincard, minimum width=1.10cm,
      font=\sffamily\fontsize{8.0}{8.7}\selectfont]
    (motionA) at (0.67,-3.46)
    {motion $A$};
\node[traincard, minimum width=1.82cm] (intervention) at (2.42,-3.46)
    {local transform\\$B=\mathcal{T}_S(A)$};
\node[traincard, minimum width=1.82cm] (alignedTarget) at (4.62,-3.46)
    {aligned target\\change / preserve};
\node[traincard, minimum width=1.38cm,
      font=\sffamily\fontsize{8.0}{8.7}\selectfont]
    (encodeTarget) at (6.62,-3.46)
    {encode\\target};
\node[traincard, minimum width=1.18cm,
      font=\sffamily\fontsize{8.0}{8.7}\selectfont]
    (addNoise) at (8.28,-3.46)
    {add\\noise};
\node[traincard, minimum width=1.48cm,
      font=\sffamily\fontsize{8.0}{8.7}\selectfont]
    (prediction) at (10.38,-3.46)
    {model\\prediction};
\node[traincard, draw=msTrain!68, fill=msTrainFill,
      text=msTrain, minimum width=2.42cm, minimum height=0.70cm]
    (routingLoss) at (13.24,-3.46)
    {\textbf{Routing losses}\\change \;\textbullet\; preserve \;\textbullet\; realism};

\draw[stripflow] (motionA.east) -- (intervention.west);
\draw[stripflow] (intervention.east) -- (alignedTarget.west);
\draw[stripflow] (alignedTarget.east) -- (encodeTarget.west);
\draw[stripflow] (encodeTarget.east) -- (addNoise.west);
\draw[stripflow] (prediction.east) -- (routingLoss.west);
\draw[trainflow] (addNoise.north) -- ++(0,0.34cm) -|
    ($(dit.south)+(-0.72cm,0)$);
\draw[trainflow] ($(dit.south)+(0.72cm,0)$) -- ++(0,-0.28cm) -|
    (prediction.north);
\draw[trainflow] (encodeTarget.south) -- ++(0,-0.27cm) -|
    (routingLoss.south);

% ---------------------------------------------------------------------------
% Minimal legend
% ---------------------------------------------------------------------------
\node[anchor=west, inner sep=0pt,
      font=\sffamily\fontsize{7.7}{8.4}\selectfont\bfseries,
      text=msInk] (legendTitle) at (0.00,-4.31) {Legend};

\node[swatch, draw=msRule, fill=msFrozen, right=0.20cm of legendTitle]
    (swFrozen) {};
\node[legendtext, right=0.07cm of swFrozen] (txFrozen) {frozen};
\node[swatch, draw=msProposed, fill=msProposedFill,
      right=0.34cm of txFrozen] (swProposed) {};
\node[legendtext, right=0.07cm of swProposed] (txProposed) {trainable / proposed};
\node[swatch, draw=msAux, fill=msAuxFill,
      right=0.34cm of txProposed] (swAux) {};
\node[legendtext, right=0.07cm of swAux] (txAux) {auxiliary};

\node[minimum width=0.48cm, minimum height=0.18cm, inner sep=0pt,
      right=0.38cm of txAux] (solidLine) {};
\draw[-{Stealth[length=2.5pt,width=2.9pt]}, draw=msMuted,
      line width=0.64pt] (solidLine.west) -- (solidLine.east);
\node[legendtext, right=0.07cm of solidLine] (txSolid) {main path};

\node[minimum width=0.48cm, minimum height=0.18cm, inner sep=0pt,
      right=0.38cm of txSolid] (dashLine) {};
\draw[-{Stealth[length=2.5pt,width=2.9pt]}, draw=msTrain,
      dashed, line width=0.64pt] (dashLine.west) -- (dashLine.east);
\node[legendtext, right=0.07cm of dashLine] (txDash) {training only};

\node[minimum width=0.92cm, minimum height=0.18cm, inner sep=0pt,
      right=0.38cm of txDash] (partsLegend) {};
\foreach \dx/\c in {-0.34/msSpine,-0.20/msHead,-0.07/msLArm,
                     0.07/msRArm,0.20/msLLeg,0.34/msRLeg}{
    \node[partdot, fill=\c] at ($(partsLegend.center)+(\dx cm,0)$) {};
}
\node[legendtext, right=0.07cm of partsLegend] {body parts};

\end{tikzpicture}%
}
\caption{\textbf{MoSAIC architecture.}
The content motion is factorized into local motion groups and a separate root-trajectory condition, while references provide part-level style features. A user-specified source map routes content or reference conditions to each anatomical region. The trainable denoiser combines a local content bottleneck, anatomical attention, and part-specific style modulation; aligned interventions provide training-only change and preservation supervision.}
\label{fig:mosaic_architecture}
\end{figure}

%% file: figures/results/aligned-intervention.tex
\begin{figure*}[t]
\centering
\includegraphics[width=\textwidth]{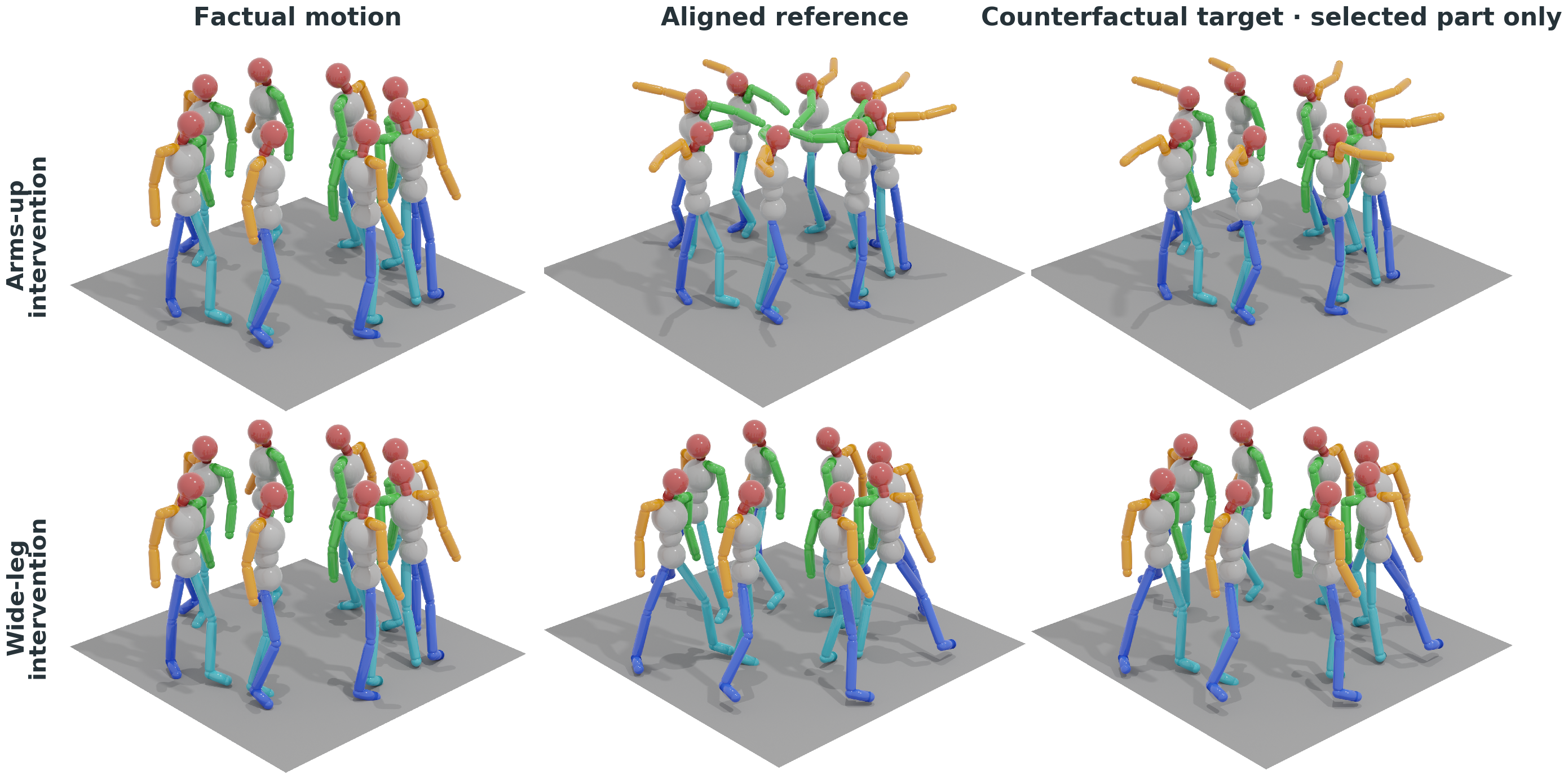}
\caption{\textbf{Aligned intervention supervision}
A controlled transformation produces a synchronized reference, while applying the same transformation only to the routed region produces the counterfactual target. Factual motion, reference, and target share action, timing, sequence length, and root trajectory, making selected-region response and preservation directly observable. Each row overlays eight ordered poses; colors identify anatomical chains.}
\label{fig:aligned_intervention}
\end{figure*}

%% file: sections/04-experiments.tex
\section{Experiments}
\label{sec:experiments}

We evaluate five questions: whether the selected model retains HumanML3D
generation quality; where it lies on the whole-body content--style frontier;
whether masked routing improves local response while limiting off-target change;
whether continuing aligned intervention supervision strengthens routed response
under matched compute; and which qualitative behaviors and limitations remain.

\subsection{Experimental Protocol}

\noindent\textbf{Model selection and inference.}
Training and aligned-pair construction use only the HumanML3D training split
\cite{guo2022generating}. We select the final model by minimum FID on the
validation split and evaluate it afterward on the disjoint test split. Unless
otherwise stated, inference uses 50-step DDIM sampling with \(\eta=0\),
zero-style classifier-free guidance, and the predeclared primary scale
\(s=2.0\). The architecture, model-selection rule, inference settings, and
evaluation manifests are fixed before final evaluation. Earlier development
was informed by previous HumanML3D and CMU observations; the reported results
are therefore development-aware.

\noindent\textbf{Whole-body evaluation.}
We follow the public MCM-LDM CMU protocol
\cite{song2024mcmldm}. Style recognition accuracy (SRA), trajectory style
inconsistency (TSI), and foot-sliding factor (FSF) are evaluated on 900 ordered
content--reference pairs. Content recognition accuracy (CRA) and Fr\'echet
motion distance (FMD) are evaluated on a separate 1,600-pair grid. Higher SRA
and CRA and lower FMD, TSI, and FSF are preferred. HumanML3D FID,
R-precision, and diversity are reported separately to verify that the
part-local training procedure does not cause gross degradation of the base
motion generator.

\noindent\textbf{Part-local evaluation.}
The primary locality benchmark contains 128 deterministically selected
HumanML3D test motions and seven aligned interventions per motion, giving 896
routed conditions. For each condition, the content-style control \(Y_0\),
whole-body route \(Y_{\mathrm{full}}\), masked route \(Y_M\), and wrong-route
control use identical initial diffusion noise. Joint errors are measured after
root alignment. If \(\mathcal C\) denotes the joints changed by the exact
counterfactual target \(X_M\), selected-target improvement is

\begin{equation}
\mathrm{STI}
=
1-
\frac{d_{\mathcal C}(Y_M,X_M)}
     {d_{\mathcal C}(Y_0,X_M)+\epsilon}.
\label{eq:sti}
\end{equation}

Preserved-region error (PRE) measures deviation from the factual motion on
unchanged joints, conditional off-target leakage (COL) measures matched-noise
change from \(Y_0\) in those regions, and exact counterfactual error (ECE)
measures whole-motion distance to \(X_M\). Higher STI and lower PRE, COL, and
ECE are preferred. The wrong-route control tests whether the response depends
on the assigned anatomical region, while a six-by-six influence matrix measures
cross-part response.

HumanML3D test values are reported over 20 fixed sampling seeds from one trained
model. For the aligned benchmark, confidence intervals use
content-motion-clustered bootstrap resampling, and route-wise comparisons use
paired Wilcoxon tests with Holm correction. Complete training settings, metric
definitions, robustness subsets, and statistical details are provided in
Appendix~\ref{app:experiment_details}.
\subsection{Generation and Whole-Body Transfer}

\input{tables/humanml-generation}

Table~\ref{tab:humanml_final} evaluates whether the final model retains
stable HumanML3D generation quality. H-FID changes from 0.2953 on validation
to \(0.3077\pm0.0032\) on the disjoint test split, while text--motion
R-precision and embedding diversity remain stable across sampling seeds. These
results indicate stable generation quality under the HumanML3D evaluator;
reference-style transfer is evaluated separately below. Model selection used
only the validation loader, and all reported test replications were generated
after the checkpoint and inference setting were frozen.

\input{tables/whole-body-cmu}

The CMU experiment verifies that the selected checkpoint retains a viable
whole-body content--style operating point. At \(s=2.0\), \mosaic obtains 50.33\% SRA and 38.13\%
CRA (Table~\ref{tab:sota}). Relative to the published MCM-LDM point, \mosaic
retains more classifier-measured content, with weaker style recognition and
higher FMD, and obtains lower FSF. These values place the selected model at a
different point on the whole-body content--style frontier. The primary
advantage of \mosaic is evaluated in the part-local setting, which these
aggregate metrics do not measure. Because the published baselines were not regenerated with matched
noise and their sample-level predictions are unavailable, the differences are
descriptive rather than paired statistical comparisons.

\subsection{Part-Local Counterfactual Evaluation}
\label{sec:partlocal_quant}

\input{tables/part-local-controls}

The primary question is not whether a reference can produce a large motion
change, but whether that change is concentrated in the requested anatomical
region. Table~\ref{tab:partlocal_primary} reveals a clear
response--preservation trade-off. Whole-body routing produces the strongest
selected-region response, but it also changes regions that the exact
counterfactual target preserves. Masked routing gives a slightly weaker
response while substantially reducing unnecessary change. It reaches an STI
of 0.198, improves exact counterfactual error by 3.56\,mm relative to the
content-style control \(Y_0\), and reduces preserved-region error and
conditional off-target leakage by 4.19\,mm and 8.20\,mm, respectively,
relative to whole-body routing.

This pattern directly supports the design goal of \mosaic. The method is not
intended to maximize reference strength independently of content preservation.
Instead, anatomical routing restricts the reference effect so that the
requested region responds while the remainder of the motion stays closer to
the content sequence. The wrong-route control further supports this
interpretation: assigning the changed condition to an unrelated region
degrades all primary metrics. The improvement therefore depends on the
correspondence between the reference cue and its assigned anatomical region,
rather than merely on providing the model with additional style information.

The paired analysis confirms that the aggregate trend is consistent across the
evaluation population. The motion-clustered 95\% confidence interval for
masked STI is \([0.178,\,0.217]\). Relative to whole-body routing, the
reductions in PRE and COL are \([3.83,\,4.55]\)\,mm and
\([7.86,\,8.54]\)\,mm, respectively. These results indicate that masking
provides a reliable improvement in preservation, rather than an isolated gain
from a small number of motions.

\input{tables/route-wise-response}

The route-wise results show how this benefit varies across anatomical
regions. Responses are strongest for the legs and spine, while
both arms show smaller but positive position-based STI
(Table~\ref{tab:partlocal_routes}). This suggests that the current
part-specific representation and conditioning pathway capture large
lower-body and torso changes more effectively than subtler upper-body cues.
The route-wise analysis therefore localizes the main remaining performance gap
to upper-body conditioning, particularly the arm pathways.

Position-based STI is undefined for the head route because the controlled head
intervention changes orientation without necessarily moving the terminal joint.
A supplemental rotation-space measure gives a head R-STI of 0.346 with a 95\%
confidence interval of \([0.289,\,0.400]\). We treat this as supporting
evidence because the metric was introduced after identifying the limitation of
the position-only endpoint.

The influence matrix provides a complementary view of locality. It places
42.4\% of the total matched-noise response on the requested-route diagonal,
compared with 16.7\% under uniform influence. Thus, the model responds
preferentially to the assigned region rather than distributing the effect
uniformly across the body. The diagonal is not dominant enough to imply
complete isolation, however. Spine edits also affect the head and arms, which
is partly expected because proximal rotations propagate through the
kinematic chain. The full matrix and route-wise uncertainty are reported in
Appendix~\ref{app:full_local_results}.

Finally, the same response pattern remains stable across three sampling seeds
and unseen intervention magnitudes. STI has a between-seed standard deviation
of 0.005 and remains positive at both \(0.75\times\) and \(1.25\times\) the
training intervention magnitude. Root-trajectory error and contact F1 also
remain close to the content-style control. Together, these results show that
the observed local response is not produced by a single sampling outcome and
that selective routing can be introduced without materially disrupting the
global trajectory or contact structure.

Overall, the evaluation supports the central claim of \mosaic: anatomically
masked routing produces a more favorable editing behavior than unrestricted
whole-body conditioning by balancing the requested regional response against
preservation of the remaining motion.

\subsection{Effect of Retaining Aligned Intervention Supervision During Continued Training}
\label{sec:aligned_supervision_ablation}

We examine whether retaining aligned intervention supervision strengthens the
response to user-routed references under matched training conditions. Both
variants start from the same previously aligned-supervised model and share the
architecture, trainable parameters, data, optimizer, training budget,
selection rule, and frozen evaluation protocol. A0 continues with factual
self-denoising only, whereas A1 retains the aligned counterfactual rows and
their associated objectives. The comparison therefore measures the effect of
\emph{continuing} aligned supervision rather than training without any prior
aligned-intervention exposure.

\begin{table*}[t]
\caption{\textbf{Matched-budget aligned-supervision continuation.}
A0 and A1 share the architecture, initialization, optimization budget,
selection rule, and evaluation protocol; only A1 retains aligned intervention
supervision. Arrows indicate preferred directions.}
\label{tab:aligned_supervision_ablation}
\centering
\footnotesize
\setlength{\tabcolsep}{3.75pt}

\textbf{(a) Part-local response and preservation}\\[2pt]
\begin{tabular}{lrrrrrr}
\toprule
Variant &
STI \(\uparrow\) &
Diag. (\%) \(\uparrow\) &
PRE (mm) \(\downarrow\) &
COL (mm) \(\downarrow\) &
ECE (mm) \(\downarrow\) &
A-FID \(\downarrow\) \\
\midrule
A0 factual continuation
& 0.1816 & 40.45 & \textbf{65.19} & \textbf{9.89}
& \textbf{77.08} & 0.3782 \\
A1 aligned continuation
& \textbf{0.1976} & \textbf{42.49} & 67.24 & 10.01
& 79.08 & \textbf{0.3768} \\
\midrule
A1--A0
& \(+0.0160\) & \(+2.03\) & \(+2.05\) & \(+0.12\)
& \(+2.00\) & \(-0.0014\) \\
\bottomrule
\end{tabular}

\vspace{4pt}
\textbf{(b) Physical and whole-body diagnostics}\\[2pt]
\begin{tabular}{lrrrrr}
\toprule
Variant &
Contact F1 \(\uparrow\) &
Sliding (m) \(\downarrow\) &
SRA (\%) \(\uparrow\) &
CRA (\%) \(\uparrow\) &
FMD \(\downarrow\) \\
\midrule
A0 factual continuation
& \textbf{0.9500} & \textbf{1.1616} & 47.78 & 37.94 & 38.58 \\
A1 aligned continuation
& 0.9472 & 1.1756 & \textbf{50.22} & \textbf{38.56}
& \textbf{36.14} \\
\midrule
A1--A0
& \(-0.0028\) & \(+0.0140\) & \(+2.44\) & \(+0.63\)
& \(-2.43\) \\
\bottomrule
\end{tabular}
\end{table*}

Continuing aligned supervision increases masked STI from 0.1816 to 0.1976 and
raises requested-route influence concentration from 40.45\% to 42.49\%
(Table~\ref{tab:aligned_supervision_ablation}). The paired confidence intervals
for both differences exclude zero. The gain is not uniform across anatomy:
response improves for the right arm and both legs, decreases for the spine,
and shows no detectable change for the left arm.

The stronger routed response introduces a modest preservation cost. PRE and
ECE increase by 2.05 and 2.00\,mm, respectively, while the 0.12\,mm change in
conditional off-target leakage has a confidence interval that includes zero.
A1 also improves the fixed CMU operating point, increasing SRA and CRA by
2.44 and 0.63 percentage points and reducing FMD by 2.43. Overall, retaining aligned supervision shifts the model toward stronger use
and greater anatomical concentration of routed reference conditions. This
improves selected-region response while introducing a small preservation cost,
consistent with a deliberate shift along the response--preservation frontier.
The result is distinct from the masked-versus-whole-body comparison in
Sec.~\ref{sec:partlocal_quant}, where masking reduces unnecessary changes
outside the requested regions.
\subsection{Qualitative Results and Guidance Behavior}
\label{sec:qualitative_routing}

\input{figures/results/part-local-transfer}

Figure~\ref{fig:part_local_transfer} compares matched inputs and diffusion noise
for arms, legs, and spine. Changing the source map changes the requested
anatomical chain while retaining the content and trajectory source. The weaker
visible arm response is consistent with the route-wise quantitative analysis.

\input{figures/results/multi-reference-composition}

Figure~\ref{fig:multi_reference_composition} assigns one reference to both arms
and another to both legs in a single denoising run; the head and spine retain
content-derived conditions. The output remains a coherent circular walk and
exhibits both routed cues. Together, Figs.~\ref{fig:part_local_transfer} and
\ref{fig:multi_reference_composition} visualize the routing behavior quantified
in Sec.~\ref{sec:partlocal_quant} and demonstrate single-pass composition from
two independently routed references.

\input{figures/results/guidance-sweep}

Increasing guidance raises SRA while reducing CRA, and FMD improves only to an
intermediate scale before worsening. Figure~\ref{fig:guidance_visual} shows the
corresponding gradual change under matched noise. No scale dominates all
metrics, so \(s=2.0\) remains the primary operating point; the complete sweep is
reported in Appendix~\ref{app:guidance}. Across the frozen sweep, SRA rises from
50.33\% to 60.78\% as CRA falls from 38.13\% to 32.75\%, while FMD reaches its
minimum at \(s=2.75\). The sweep therefore describes an inference-time frontier
rather than identifying a replacement primary result.

\subsection{Limitations}

The evaluation has three principal boundaries. First, the primary model and
the matched continuation study each use one training seed, so the reported
sampling-seed analysis varies diffusion noise rather than independent training
runs. Second, the aligned benchmark uses controlled interventions because
naturally paired part-local targets are unavailable; it therefore evaluates
selective response and preservation under known counterfactuals rather than
the full diversity of natural motion styles. Third, the study does not include
matched local retraining of prior methods, a human perceptual study, or a
population-level evaluation of multi-reference composition. Accordingly, our
claims concern the effect of aligned supervision and anatomical masking within
\mosaic rather than overall superiority to prior systems. The route-wise
analysis also identifies weaker arm response and incomplete cross-part
isolation as the main technical limitations of the current model. Earlier
development was informed by HumanML3D test and CMU observations, and the final
reported evaluation should therefore be interpreted as development-aware.

\FloatBarrier

%% file: tables/humanml-generation.tex
\begin{table}[t]
\centering
\caption{Validation selection and disjoint HumanML3D test generation quality.
Test values are means \(\pm\) 95\% confidence intervals over 20 sampling
seeds from the validation-selected model.}
\label{tab:humanml_final}
\begingroup
\footnotesize
\setlength{\tabcolsep}{2.5pt}
\renewcommand{\arraystretch}{1.08}
\begin{tabular*}{\columnwidth}{@{\extracolsep{\fill}}lrrrrr@{}}
\toprule
Split & H-FID $\downarrow$ & R@1 (\%) $\uparrow$ & R@2 (\%) $\uparrow$ &
R@3 (\%) $\uparrow$ & Div.\\
\midrule
Validation & 0.2953 & 47.21 & 66.56 & 77.26 & 9.205\\
Test & $0.3077{\pm}0.0032$ & $45.32{\pm}0.23$ &
$64.35{\pm}0.28$ & $74.70{\pm}0.27$ & $9.174{\pm}0.071$\\
\bottomrule
\end{tabular*}
\endgroup
\end{table}

%% file: tables/whole-body-cmu.tex
\begin{table}[t]
\centering
\caption{Whole-body transfer on the MCM-LDM CMU protocol. Baselines are
published values; \mosaic uses the frozen primary setting. The comparison is
not a seed-matched reproduction.}
\label{tab:sota}
\begingroup
\footnotesize
\setlength{\tabcolsep}{2.0pt}
\renewcommand{\arraystretch}{1.08}
\begin{tabular*}{\columnwidth}{@{\extracolsep{\fill}}lrrrrr@{}}
\toprule
Method & SRA (\%) $\uparrow$ & CRA (\%) $\uparrow$ & FMD $\downarrow$ &
TSI $\downarrow$ & FSF $\downarrow$\\
\midrule
1DConv+AdaIN~\cite{aberman2020unpaired} & 57.00 & 31.18 & 42.68 & 0.2200 & 2.0500\\
STGCN+AdaIN~\cite{park2021diverse} & 17.66 & \textbf{60.43} & 129.44 & \textbf{0.1100} & 0.9300\\
Motion Puzzle~\cite{jang2022motionpuzzle} & 46.33 & 26.31 & 113.31 & 0.2200 & 2.4300\\
MCM-LDM~\cite{song2024mcmldm} & \textbf{58.00} & 35.75 & \textbf{27.69} & 0.4000 & 1.2800\\
\midrule
\mosaic & 50.33 & 38.13 & 39.88 & 0.5020 & \textbf{0.8285}\\
\bottomrule
\end{tabular*}
\endgroup
\end{table}

%% file: tables/part-local-controls.tex
\begin{table}[t]
\centering
\caption{Aligned evaluation on 128 test motions and 896 conditions. Errors
are root-aligned millimeters. Position STI excludes 128 head conditions,
whose intervention changes terminal-joint orientation only.}
\label{tab:partlocal_primary}
\begingroup
\footnotesize
\setlength{\tabcolsep}{3.0pt}
\renewcommand{\arraystretch}{1.08}
\begin{tabular*}{\columnwidth}{@{\extracolsep{\fill}}lrrrr@{}}
\toprule
Control & STI $\uparrow$ & PRE $\downarrow$ & COL $\downarrow$ & ECE $\downarrow$\\
\midrule
Content style $Y_0$ & 0.000 & \textbf{64.84} & \textbf{0.00} & 81.80\\
Whole body $Y_{\mathrm{full}}$ & \textbf{0.215} & 70.64 & 18.08 & 81.42\\
\textbf{Masked $Y_M$} & 0.198 & 66.45 & 9.88 & \textbf{78.24}\\
Wrong route & -0.290 & 106.03 & 84.65 & 127.11\\
\bottomrule
\end{tabular*}
\endgroup
\end{table}

%% file: tables/route-wise-response.tex
\begin{table}[t]
\centering
\caption{Masked response by requested route over 128 motions. P-STI uses
joint positions; supplemental R-STI uses articulation-joint rotation because
the terminal head intervention is position-invisible.}
\label{tab:partlocal_routes}
\begingroup
\footnotesize
\setlength{\tabcolsep}{3.5pt}
\renewcommand{\arraystretch}{1.06}
\begin{tabular*}{\columnwidth}{@{\extracolsep{\fill}}lrrr@{}}
\toprule
Route & P-STI $\uparrow$ & R-STI $\uparrow$ & COL (mm) $\downarrow$\\
\midrule
Spine & 0.142 & 0.179 & 11.99\\
Head & -- & \textbf{0.346} & 7.02\\
Left arm & 0.090 & 0.190 & \textbf{6.76}\\
Right arm & 0.065 & 0.235 & 9.70\\
Left leg & 0.335 & 0.103 & 8.75\\
Right leg & \textbf{0.413} & 0.163 & 12.98\\
\bottomrule
\end{tabular*}
\endgroup
\end{table}

%% file: figures/results/part-local-transfer.tex
\begin{figure*}[t]
\centering
\includegraphics[width=\textwidth]{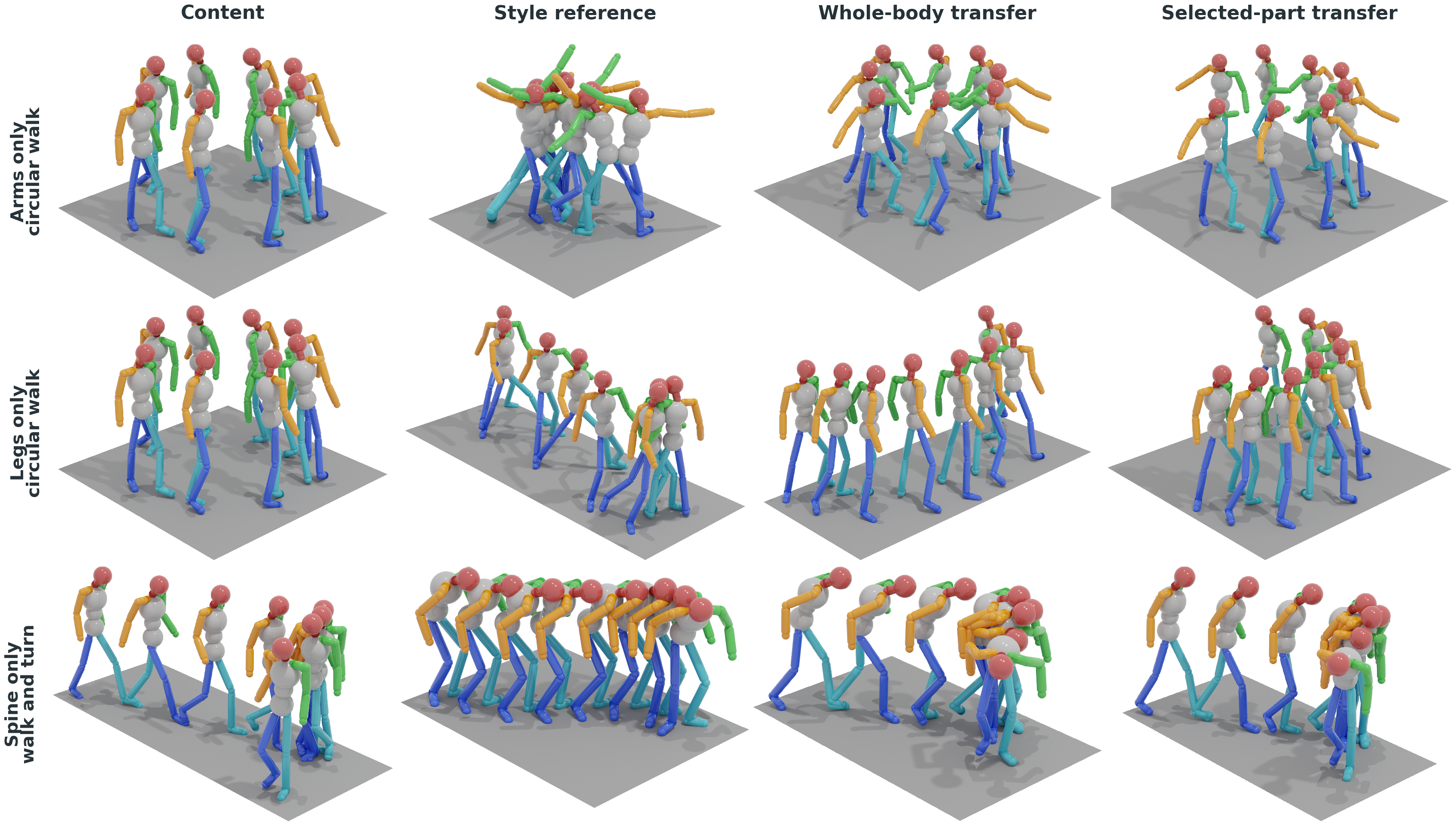}
\caption{\textbf{Part-local routing.}
Rows assign hands-high style to the arms, open-leg style to the legs, and old-motion style to the spine. Columns show the content motion, style reference, whole-body transfer, and selected-part transfer. Content, reference, and initial diffusion noise are matched within each row at the primary setting. The examples visualize the route-wise behavior quantified in Tables 3 and 4.}
\label{fig:part_local_transfer}
\end{figure*}

%% file: figures/results/multi-reference-composition.tex
\begin{figure*}[t]
\centering
\includegraphics[width=\textwidth]{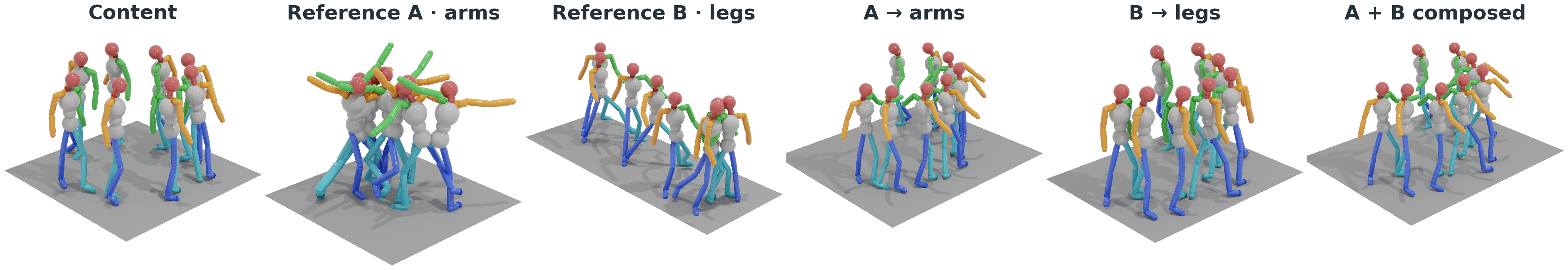}
\caption{\textbf{Two-reference source assignment.}
Circular-walk content is combined with hands-high and open-leg references. The single-route controls assign the first reference to the arms and the second to the legs, while the composed route applies both assignments in one denoising run and retains content-derived conditions elsewhere. All generated panels use matched initial noise.}
\label{fig:multi_reference_composition}
\end{figure*}

%% file: figures/results/guidance-sweep.tex
\begin{figure*}[t]
\centering
\includegraphics[width=\textwidth]{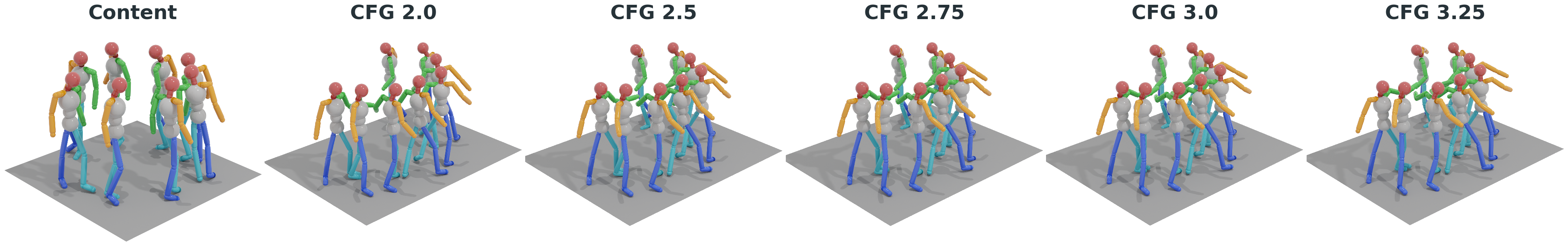}
\caption{\textbf{Guidance behavior.}
The content motion, two routed references, and initial noise are held fixed while classifier-free guidance increases from s=2.0 to s=3.25. The routed response changes gradually across the sweep. The predeclared primary setting is s=2.0; the remaining panels are frozen-checkpoint sensitivity diagnostics.}
\label{fig:guidance_visual}
\end{figure*}

%% file: sections/05-conclusion.tex
\section{Conclusion}
\label{sec:conclusion}

We presented \mosaic, a latent diffusion framework for part-local
reference-conditioned motion style transfer. The central challenge is that
self-reconstruction does not reveal the counterfactual response that should
follow when a single regional reference is replaced. \mosaic addresses this
problem with aligned intervention supervision, which constructs synchronized
references and exact counterfactual targets for both selected-region response
and preservation. A user-specified source map, part-factorized conditioning, a
separate root-trajectory pathway, and anatomically constrained denoising
translate this supervision into controllable regional routing.

On a frozen benchmark of 128 held-out motions and 896 routed conditions,
masked routing provides a more favorable response--preservation balance than
whole-body routing. It retains a positive selected-target response while
reducing preserved-region error by 4.19\,mm and matched-noise off-target
leakage by 8.20\,mm. The response is preferentially concentrated in the
requested anatomical region, and the wrong-route control confirms that this
behavior depends on the correspondence between the reference cue and its
assigned body part. A matched-budget continuation study further shows that
retaining aligned intervention supervision shifts the model toward stronger
and more anatomically concentrated use of routed references.

The present evidence concerns controlled aligned interventions and six
sequence-wide body regions; natural-reference benchmarks, population-level
multi-reference evaluation, and independent training seeds remain important
directions for future work. Within this scope, the results establish aligned
intervention supervision as an effective mechanism for learning selective
local response when naturally paired part-local targets are unavailable.

%% file: appendices/experiment-details.tex
\section{Experimental Details}
\label{app:experiment_details}

\subsection{Full Training and Inference Configuration}

HumanML3D is represented by 22 joints and 263 rotation-invariant features at
20\,fps.  After the released mirroring augmentation, the local split files
contain 23,384 training, 1,460 validation, and 4,384 test identifiers; these
are processed identifiers rather than unique source motions.  Their pairwise
overlap is zero.  Training motions are restricted to 40--196 frames.

The selected run was continued from the earlier temporal-injection checkpoint
and trained through epoch 200 with early stopping disabled.  Six NVIDIA L40S
GPUs processed 64 motions each, giving global batch size 384.  Of 326.01\,M
parameters, 27.42\,M were trainable.  AdamW used a fixed learning rate of
\(10^{-5}\) and seed 1234.  After a 20-epoch curriculum, batches contained
75\% factual self-denoising rows and 25\% aligned same-motion intervention
rows; no arbitrary unpaired row was used.  Factual rows sampled all 1,000
diffusion steps, while intervention rows sampled \(t\in[0,400)\).  The
scaled-linear DDPM schedule used
\(\beta_1=8.5{\times}10^{-4}\) and \(\beta_{1000}=0.012\).

The diffusion objective weight was 1.0.  Weights for selected-part response,
unselected-part preservation, style-code matching, source margin, root,
contact, velocity, decorrelation, and descriptor matching were respectively
0.025, 0.05, 0.0125, 0.0125, 0.05, 0.0125, 0.025, 0.0025, and 0.025.
Inference used 50 DDIM steps, \(\eta=0\), zero-style guidance, and primary
scale \(s=2.0\).  HumanML3D test evaluation used seeds 1234--1253.  CMU and
the primary aligned evaluation used seed 1234.

The CMU style grid contains \(30\times30=900\) ordered pairs from six style
classes; the content/realism grid contains \(40\times40=1,600\) pairs from
eight content classes.  The checkpoint, sampler, two-worker mapping, ordered
pair lists, evaluators, and asset digests were frozen in evaluation manifests
before final generation.  The aligned suite fixes 128 test motions by
identifier hash and applies seven conditions per motion: two spine, one head,
two arm, and two leg interventions.  The first 32 motions define the
sampling-seed subset and the first 64 define the intervention-magnitude
subset.

\subsection{Metric Definitions}

\input{tables/metric-definitions}

The changed set contains joints whose factual-to-counterfactual displacement
exceeds 1\,mm; its complement defines preserved joints.  The primary
position-only STI is undefined for the 128 head-nod conditions because a
terminal-joint rotation need not displace that joint.  R-STI is a post-hoc
supplement based on SO(3) geodesic error at the intervention articulation
joint.  It does not replace position STI.

HumanML3D confidence intervals summarize 20 fixed generation seeds from one
training run.  The aligned intervals use 10,000 bootstrap resamples clustered
by content motion.  Route-wise comparisons use two-sided paired Wilcoxon
tests with Holm correction over the six routes and paired standardized mean
differences.  CMU uses one fixed generation seed over each full grid.  The
external aggregate reports do not contain per-example classifier decisions,
precluding paired bootstrap comparisons with published methods.

\subsection{Guidance Sensitivity}
\label{app:guidance}

\input{tables/guidance-sensitivity}

Increasing \(s\) from 2.00 to 3.25 raises SRA by 10.44 percentage points and
reduces CRA by 5.38 points.  FMD reaches its minimum of 37.49 at \(s=2.75\)
and then worsens, whereas TSI increases monotonically and FSF decreases.  At
\(s=2.50\), SRA is 0.11 points above the published MCM-LDM result, but CRA is
0.63 points lower and FMD remains 10.15 higher.  No sensitivity row dominates
all metrics or replaces the predeclared \(s=2.0\) result.

\subsection{Sampling-Seed and Intervention-Magnitude Robustness}

\input{tables/part-local-robustness}

Response remains positive under both unseen intervention magnitudes.  The
stronger intervention increases STI but also COL and ECE, exposing the
response--preservation trade-off.  The seed rows vary only diffusion sampling
noise for one trained model and therefore do not establish training-seed
robustness.

\subsection{Auxiliary Trajectory, Contact, and Sliding Diagnostics}
\label{app:auxiliary}

\input{tables/auxiliary-diagnostics}

Masked routing keeps trajectory error and contact F1 close to the
content-style control and improves A-FID from 0.3984 to 0.3780.  Whole-body
routing obtains lower A-FID and sliding but causes more off-target change in
Table~\ref{tab:partlocal_primary}.  The wrong route degrades every diagnostic,
showing why global distribution quality cannot replace explicit locality
measurement.

\subsection{Full Influence and Uncertainty Results}
\label{app:full_local_results}

\input{tables/influence-matrix}

\input{tables/route-wise-uncertainty}

All five position-identifiable routes have positive intervals and remain
detectable after family-wise correction.  The arm responses and paired effect
sizes are nevertheless substantially smaller than those of the spine and
legs.  For the supplemental orientation endpoint, head R-STI is 0.346 with a
motion-bootstrap 95\% interval of [0.289, 0.400].  The influence diagonal
contains 42.4\% of total response; large spine-to-head and spine-to-arm terms
show that anatomical isolation remains incomplete.

%% file: tables/metric-definitions.tex
\begin{table}[t]
\centering
\caption{Definitions, evaluation populations, and preferred directions of
reported metrics. Feature distances are measured in the corresponding learned
evaluator space.}
\label{tab:metrics}
\begingroup
\footnotesize
\setlength{\tabcolsep}{3.0pt}
\renewcommand{\arraystretch}{1.12}
\begin{tabularx}{\columnwidth}{@{}l>{\raggedright\arraybackslash}Xcc@{}}
\toprule
Metric & Definition & Unit / \(N\) & Direction\\
\midrule
SRA & Top-1 target-style classification of transferred motion
    & \% / 900 & \(\uparrow\)\\
CRA & Top-1 source-content classification of transferred motion
    & \% / 1,600 & \(\uparrow\)\\
FMD & Fr\'echet distance between real and generated content-classifier features
    & feature distance / 1,600 & \(\downarrow\)\\
TSI & Mean per-frame horizontal root-path error to content
    & m / 900 & \(\downarrow\)\\
FSF & Two-foot horizontal displacement accumulated below 5\,cm
    & m per motion / 900 & \(\downarrow\)\\
H-FID & Fr\'echet distance in the HumanML3D text--motion evaluator
    & feature distance & \(\downarrow\)\\
R@1 & Top-1 HumanML3D text--motion retrieval precision
    & \% & \(\uparrow\)\\
Div. & Mean pairwise distance between generated motion embeddings
    & feature distance & target-matched\\
STI & Changed-joint target-error reduction relative to matched \(Y_0\)
    & ratio / 768 & \(\uparrow\)\\
PRE & Error to factual motion on exact unchanged joints
    & mm / 896 & \(\downarrow\)\\
COL & Matched-noise change from \(Y_0\) on exact unchanged joints
    & mm / 896 & \(\downarrow\)\\
ECE & Whole-motion error to the exact aligned counterfactual
    & mm / 896 & \(\downarrow\)\\
\bottomrule
\end{tabularx}
\endgroup
\end{table}

%% file: tables/guidance-sensitivity.tex
\begin{table}[t]
\centering
\caption{Frozen-checkpoint CMU guidance sweep. The dagger marks the
predeclared primary setting; all other rows are secondary sensitivity
measurements on identical ordered grids.}
\label{tab:cfg}
\begingroup
\footnotesize
\setlength{\tabcolsep}{2.4pt}
\renewcommand{\arraystretch}{1.06}
\begin{tabular*}{\columnwidth}{@{\extracolsep{\fill}}crrrrr@{}}
\toprule
Scale & SRA (\%) $\uparrow$ & CRA (\%) $\uparrow$ & FMD $\downarrow$ &
TSI $\downarrow$ & FSF $\downarrow$\\
\midrule
2.00$^\dagger$ & 50.33 & 38.13 & 39.88 & 0.5020 & 0.8285\\
2.25 & 55.11 & 36.44 & 38.62 & 0.5053 & 0.8214\\
2.50 & 58.11 & 35.13 & 37.84 & 0.5086 & 0.8151\\
2.75 & 59.89 & 34.13 & 37.49 & 0.5116 & 0.8095\\
3.00 & 60.67 & 33.56 & 37.56 & 0.5144 & 0.8087\\
3.25 & 60.78 & 32.75 & 38.00 & 0.5169 & 0.8071\\
\bottomrule
\end{tabular*}
\endgroup
\end{table}

%% file: tables/part-local-robustness.tex
\begin{table}[t]
\centering
\caption{Masked-routing robustness subsets. Seed rows use 32 motions and all
seven conditions; magnitude rows use 64 motions. Only sampling noise or
intervention strength changes.}
\label{tab:partlocal_robustness}
\begingroup
\footnotesize
\setlength{\tabcolsep}{2.8pt}
\renewcommand{\arraystretch}{1.06}
\begin{tabular*}{\columnwidth}{@{\extracolsep{\fill}}lrrrr@{}}
\toprule
Variant & STI $\uparrow$ & PRE $\downarrow$ & COL $\downarrow$ & ECE $\downarrow$\\
\midrule
Seed 1234 & 0.230 & 54.46 & 9.10 & 66.80\\
Seed 1235 & 0.224 & 54.90 & 8.98 & 67.93\\
Seed 1236 & 0.220 & 57.15 & 8.92 & 69.31\\
\quad between-seed SD & 0.005 & 1.44 & 0.09 & 1.26\\
\midrule
\(0.75\times\) magnitude & 0.187 & 58.65 & 7.29 & 68.11\\
\(1.25\times\) magnitude & 0.228 & 60.02 & 11.86 & 74.91\\
\bottomrule
\end{tabular*}
\endgroup
\end{table}

%% file: tables/auxiliary-diagnostics.tex
\begin{table}[t]
\centering
\caption{Auxiliary aligned diagnostics at the primary setting. A-FID uses
HumanML3D evaluator features against exact targets; the exact target row is
an oracle reference, not model output.}
\label{tab:aligned_auxiliary}
\begingroup
\footnotesize
\setlength{\tabcolsep}{2.8pt}
\renewcommand{\arraystretch}{1.06}
\begin{tabular*}{\columnwidth}{@{\extracolsep{\fill}}lrrrr@{}}
\toprule
Control & A-FID $\downarrow$ & Traj. (m) $\downarrow$ &
Contact F1 $\uparrow$ & Slide (m) $\downarrow$\\
\midrule
Content style $Y_0$ & 0.3984 & 0.3342 & 0.9485 & 1.1535\\
Whole body $Y_{\mathrm{full}}$ & 0.3592 & 0.3358 & 0.9473 & 1.1236\\
\textbf{Masked $Y_M$} & 0.3780 & 0.3353 & 0.9477 & 1.1450\\
Wrong route & 3.6021 & 0.5285 & 0.9132 & 1.6852\\
Exact target $X_M$ & 0.0000 & 0.0000 & 1.0000 & 0.8634\\
\bottomrule
\end{tabular*}
\endgroup
\end{table}

%% file: tables/influence-matrix.tex
\begin{table}[t]
\centering
\caption{Matched-noise influence matrix in root-aligned millimeters. Rows are
requested routes and columns are measured regions; bold entries form the
requested-route diagonal.}
\label{tab:partlocal_influence}
\begingroup
\scriptsize
\setlength{\tabcolsep}{1.7pt}
\renewcommand{\arraystretch}{1.05}
\begin{tabular*}{\columnwidth}{@{\extracolsep{\fill}}lrrrrrr@{}}
\toprule
Request & Sp. & Hd. & L-A & R-A & L-L & R-L\\
\midrule
Spine & \textbf{17.72} & 39.86 & 28.73 & 31.93 & 13.77 & 13.42\\
Head & 2.07 & \textbf{34.71} & 7.54 & 6.35 & 6.60 & 6.84\\
Left arm & 2.68 & 6.35 & \textbf{27.35} & 8.54 & 5.97 & 5.27\\
Right arm & 4.22 & 9.75 & 12.18 & \textbf{30.82} & 8.56 & 8.91\\
Left leg & 2.08 & 5.03 & 7.31 & 7.37 & \textbf{75.78} & 11.84\\
Right leg & 2.81 & 6.78 & 10.47 & 10.07 & 16.55 & \textbf{117.56}\\
\bottomrule
\end{tabular*}
\endgroup
\end{table}

%% file: tables/route-wise-uncertainty.tex
\begin{table}[t]
\centering
\caption{Route-wise position-STI uncertainty. Intervals use
motion-clustered bootstrap resampling; \(p_{\mathrm H}\) is the
Holm-corrected paired Wilcoxon result against \(Y_0\).}
\label{tab:partlocal_significance}
\begingroup
\footnotesize
\setlength{\tabcolsep}{2.4pt}
\renewcommand{\arraystretch}{1.06}
\begin{tabular*}{\columnwidth}{@{\extracolsep{\fill}}lrrr@{}}
\toprule
Route & P-STI [95\% CI] $\uparrow$ & \(p_{\mathrm H}\) & \(d_z\)\\
\midrule
Spine & 0.142 [0.128, 0.155] & \(2.81{\times}10^{-21}\) & 1.817\\
Left arm & 0.090 [0.061, 0.120] & \(1.40{\times}10^{-6}\) & 0.527\\
Right arm & 0.065 [0.033, 0.096] & \(7.21{\times}10^{-5}\) & 0.353\\
Left leg & 0.335 [0.297, 0.372] & \(5.52{\times}10^{-21}\) & 1.541\\
Right leg & 0.413 [0.366, 0.459] & \(2.28{\times}10^{-20}\) & 1.532\\
\bottomrule
\end{tabular*}
\endgroup
\end{table}

%% file: backmatter/declarations.tex
\section*{Declarations}

% PROVISIONAL DECLARATIONS: verify every statement and adapt author roles,
% funding, conflicts, and availability terms before submission.
\subsection*{Funding}
This work was partially supported by the U.S. National Science Foundation
under Award Nos. 2211785 and 2316240. Any opinions, findings, conclusions, or
recommendations expressed in this material are those of the authors and do not
necessarily reflect the views of the U.S. National Science Foundation.

\subsection*{Competing interests}
The authors have no relevant financial or non-financial interests to disclose.

\subsection*{Author contributions}
The first author conceived the study, developed the method, implemented the
system, conducted the experiments, analyzed the results, and prepared the
initial manuscript. The second author supervised the research, contributed to
the study design and interpretation of the results, and reviewed and revised
the manuscript. All authors read and approved the final manuscript.

\subsection*{Data availability}
This study uses the publicly available HumanML3D and CMU motion-capture
datasets. The datasets are available from their respective providers, subject
to their original terms and licenses. No new dataset was collected for this
study.

\subsection*{Code availability}
The source code, evaluation scripts, and configurations are publicly available
for research use at \url{https://github.com/UTSA-VIRLab/MoSAIC}.

\subsection*{Ethics approval}
Not applicable. This study uses previously published, publicly available
motion-capture datasets and did not recruit new human participants.

\subsection*{Consent to participate}
Not applicable.

\subsection*{Consent for publication}
Not applicable.